\documentclass[11pt]{article}

\usepackage[margin=1in]{geometry}
\usepackage[T1]{fontenc}    

\usepackage{graphicx}
\usepackage{booktabs}
\usepackage{multirow}
\usepackage{wrapfig}

\usepackage{enumitem}

\usepackage{listings}

\lstdefinelanguage{json}{
    basicstyle=\ttfamily\footnotesize,
    string=[s]{"}{"},
    stringstyle=\ttfamily,
    comment=[l]{//},
    commentstyle=\itshape,
    morecomment=[s]{/*}{*/},
    keywords={true,false,null},
    keywordstyle=\bfseries,
    showstringspaces=false,
    breaklines=true,
    frame=lines,
    framesep=2mm,
    tabsize=2,
    columns=fullflexible,
    keepspaces=true
}
\lstdefinestyle{markdown}{
    basicstyle=\ttfamily\footnotesize,
    breaklines=true,
    breakatwhitespace=false,
    columns=fullflexible,
    keepspaces=true,
    showstringspaces=false,
    frame=lines,
    framesep=2mm,
    tabsize=2,
    linewidth=\linewidth
}
\usepackage[numbers]{natbib}

\usepackage{etoolbox}
\usepackage[dvipsnames]{xcolor}
\usepackage{xspace}

\usepackage[accsupp]{axessibility}

\usepackage[
    pagebackref,
    breaklinks,
    colorlinks
]{hyperref}

\AtEndPreamble{
\usepackage[capitalize]{cleveref}

\crefname{section}{Section}{Sections}
\Crefname{section}{Section}{Sections}

\crefname{table}{Table}{Tables}
\Crefname{table}{Table}{Tables}

\crefname{figure}{Figure}{Figures}
\Crefname{figure}{Figure}{Figures}
}

\makeatletter

\DeclareRobustCommand{\onedot}{%
    \futurelet\@let@token\@onedot
}

\def\@onedot{%
    \ifx\@let@token.
    \else
        .\null
    \fi
    \xspace
}

\newcommand{\eg}{\emph{e.g}\onedot}

\newcommand{\vs}{\emph{vs}\onedot}

\makeatother

\begin{document}

\title{
\Large \bfseries
FreeStory: Training-Free Character Consistency \\ for Free-Form Visual Storytelling
\rule{\textwidth}{0.7pt}
}
\author{
  Sibo Dong \quad
  Ismail Shaheen \quad
  Sarah Adel Bargal \\
  Department of Computer Science, Georgetown University \\
  \texttt{\{sd1242, ias68, sarah.bargal\}@georgetown.edu}
}
\date{} 

\maketitle

\begin{abstract}
Visual storytelling aims to generate image sequences that are both aligned with narrative prompts and consistent in character appearance across images.
Recent training-free methods improve character consistency by reusing attention features, but rely on structured prompts where full character descriptions are repeated in every prompt.
This assumption simplifies the task but deviates from natural storytelling, where characters are typically introduced once and later referred to using pronouns or type-based expressions.
We propose \textbf{FreeStory}, a training-free framework that reformulates character consistency under free-form prompts as entity-grounded feature reuse.
Our method associates reference mentions with their corresponding character descriptions and combines dynamic character masks, correspondence-aware feature matching, key-value injection, and query blending to preserve identity while retaining generation diversity.
We also introduce \textbf{FreeStoryBench}, a benchmark for this setting that includes both single- and multi-character stories.
Experiments show that FreeStory achieves state-of-the-art performance among training-free methods on structured benchmarks and stronger overall consistency over baselines under free-form prompts.
\end{abstract}

\section{Introduction}

Visual storytelling aims to generate a sequence of images that faithfully illustrate a narrative while maintaining character consistency across images.
Recent advances in large-scale diffusion models \cite{sd,sdxl,flux} have significantly improved text-to-image generation quality.
However, generating consistent characters across multiple images remains challenging, as diffusion models typically generate each image independently and lack an explicit mechanism to preserve identity over time.
Alternatively, 
personalization or image editing methods, such as IP-Adapter~\cite{ip-adapter}, and Flux-Kontext~\cite{flux1kontext} condition on reference images to preserve visual appearance effectively. However, they often exhibit reduced diversity in pose, expression, and scene composition, as strong conditioning on reference images may limit the flexibility of generation.
As a result, they are less suitable for open-ended visual storytelling, where characters are expected to appear consistently while still adapting naturally to evolving narrative contexts.

To address this, recent visual storytelling methods propose training-free feature reuse techniques that propagate character information from a reference image to subsequent target images~\cite{consistory,zhou2024storydiffusion,onepromptonestory}.
These methods achieve promising consistency by reusing the attention features associated with character description in the prompt.
However, they rely on a strict structure in which every prompt must explicitly include the full character description.
For example, the ConsiStory+ dataset \cite{onepromptonestory} uses a structured prompt format where each prompt follows a fixed \emph{``character + activity''} structure, with the character description appearing at the beginning of every prompt.
This strict format simplifies character grounding and allows feature reuse methods to directly utilize character description.

However, this assumption deviates significantly from realistic storytelling.
In natural narratives, characters are typically introduced once using a detailed description in the first prompt, and referred to using reference mentions in subsequent prompts, such as pronouns (e.g., \textit{he}, \textit{she}) or type-based expressions (e.g., \textit{the man}, \textit{the dog}).
Moreover, realistic stories often involve multiple characters interacting with each other.
For example, a character-defining prompt may introduce both \textit{a young woman wearing a yellow shirt} and \textit{a tall man with curly hair}, while subsequent referring prompts may use reference mentions such as \textit{the woman}, \textit{he}, or even \textit{they}.
Under such free-form prompts, the diffusion model no longer receives explicit appearance information in referring prompts, making it difficult to maintain consistent character when generating target images.
This challenge is further amplified in the multi-character setting, where the model must not only preserve identity but also correctly associate each reference mention with the corresponding character.
Moreover, existing methods cannot directly operate in this setting, as they rely on known description spans in every prompt and are primarily evaluated on structured datasets that focus on single-character stories.
Thus, character consistency degrades significantly when description spans are not explicitly repeated.

\begin{figure}[tb]
    \centering
    \includegraphics[width=\linewidth]{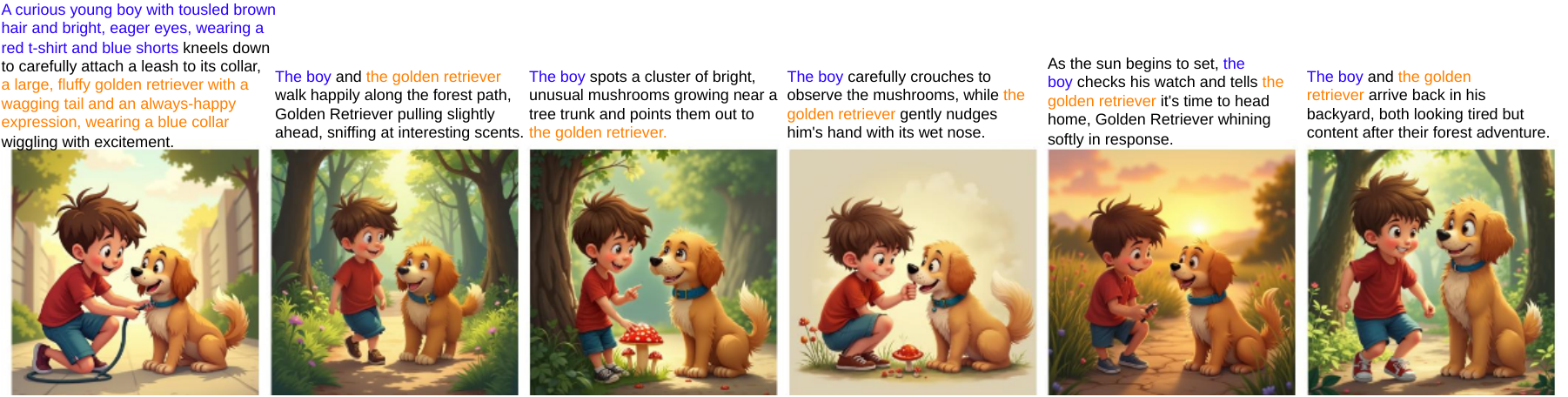}
    \caption{
\textbf{Multi-character free-form story generated by FreeStory.} 
Characters are introduced once with full descriptions and later referred to using shorter mentions (e.g., \textcolor{blue}{boy}, \textcolor{orange}{golden retriever}). 
}
    \label{fig:demo}
\end{figure}

In this work, we identify this limitation and introduce \emph{free-form visual storytelling}, where each character is introduced only once in a character-defining prompt, and subsequent referring prompts contain only reference mentions.
Our formulation naturally supports both single- and multi-character stories, where multiple characters may appear, disappear, and interact across prompts.
This setting is more realistic and substantially more challenging, as it requires maintaining consistent identity for multiple characters across reference and target images despite implicit linguistic references. \Cref{fig:demo} shows an example of such a multi-character free-form story generation.


To address free-form visual storytelling, we propose \textbf{FreeStory}, a training-free framework that reformulates consistent visual storytelling as \emph{entity-grounded feature reuse}.
Our key insight is that attention features should be transferred according to entity-level semantic correspondence rather than repeated textual descriptions or fixed token positions.
Given a reference mention in a referring prompt, an entity grounding module first associates it with the corresponding character-description span in the character-defining prompt.
The contribution of this module lies in introducing discourse-level entity grounding as the interface between free-form language and diffusion feature reuse, rather than in the particular grounding model used to instantiate it.

Entity grounding alone, however, does not make conventional feature injection directly applicable to free-form prompts.
FreeStory therefore introduces several coupled mechanisms that adapt attention reuse to this setting.
During reference generation, it dynamically extracts entity-aware attention masks and attention features associated with each character description.
During target generation, it establishes spatial correspondence between the reference and target images based on attention similarity, allowing features to be reused only at matched character regions.
It then injects reference key-value features to preserve character appearance and blends query features to retain prompt-specific variation.
Together, entity grounding, dynamic mask extraction, correspondence matching, and query blending address the linguistic alignment, spatial localization, and diversity challenges that arise specifically under free-form prompts.
This enables consistent generation without model retraining or manually specified character correspondences.

To support this new setting, we further introduce \textbf{FreeStoryBench}, a free-form visual storytelling benchmark that removes the restrictive prompt templates used in prior datasets and supports both single- and multi-character stories.
Unlike ConsiStory+~\cite{onepromptonestory}, where every prompt explicitly contains a description, FreeStoryBench allows natural narrative conventions where only the character-defining prompt contains description spans, and referring prompts contain only reference mentions.
It also provides correspondence annotations for evaluating automatic entity grounding while preserving a fully automatic test setting.

Extensive experiments demonstrate our method achieves state-of-the-art performance among training-free methods on ConsiStory+ under structured prompts.
On FreeStoryBench, it consistently improves character identity preservation over the underlying diffusion model and compared storytelling baselines in both single- and multi-character settings.
Furthermore, automatic entity grounding performs comparably to oracle ground-truth grounding, demonstrating that the proposed framework can recover the correspondences required for feature reuse without manual annotation. 

In summary, our contributions are as follows:
\begin{itemize}
\item We introduce the problem of free-form visual storytelling, moving beyond restrictive prompt templates that repeat complete character descriptions in every prompt.
\item We propose FreeStory, a training-free framework that combines entity grounding with dynamic character-mask extraction, correspondence-aware feature matching, key-value injection, and query blending to address challenges of free-form prompts.
\item We construct FreeStoryBench, a free-form storytelling benchmark that removes unrealistic prompt constraints, supports both single- and multi-character settings, and better reflects natural narrative structure.
\item Extensive experiments demonstrate state-of-the-art performance under structured prompts and stronger overall character consistency than the compared baselines under free-form prompts.
\end{itemize}




\section{Related Works}
\subsection{Visual Storytelling}

Recent progress in text-to-image (T2I) generation has been driven by diffusion-based models \cite{sd, sdxl, sd3, flux}, which demonstrate strong performance in generating high-quality images aligned with textual prompts.

Different from conventional T2I generation that produces a single image independently, visual storytelling aims to generate a sequence of images forming a coherent storyline, where each image corresponds to a segment of a text sequence. This task requires maintaining consistency in characters, objects, and visual style across images, while ensuring faithfulness to each associated prompt.
Many existing methods address visual storytelling through training-based approaches. These methods typically fine-tune diffusion models on paired image-text datasets and introduce additional mechanisms to model cross-frame dependencies, such as history memory \cite{arldm, makeastory, acmvsg, causalstory, storygen, zheng2024temporalstory, yang2024seedstory, vista}, layout control \cite{storyimager, talecrafter}, and character-aware representations \cite{storygptv, cogcartoon, storynizor, storymaker, thechosenone, wang2024oneactor}. During generation, images are commonly synthesized in an auto-regressive manner, where previously generated text-image pairs are used as additional conditions for subsequent frames.

Despite their effectiveness, training-based methods have inherent limitations. Integrating additional conditioning signals often requires architectural modifications, which in turn necessitates extensive training and large-scale datasets to achieve high-quality and consistent image generation. These limitations motivate the exploration of training-free methods that leverage pretrained diffusion models to perform visual storytelling without additional training.

\subsection{Training-free Visual Storytelling}

Training-free visual storytelling methods aim to improve character consistency across frames without introducing additional training or modifying the model architecture. A common strategy is to repeatedly leverage shared character information across story prompts, allowing pretrained diffusion models to generate consistent subjects in a purely inference-time manner.

OnePromptOneStory (1P1S)~\cite{onepromptonestory} achieves character consistency by concatenating all prompts into a single extended representation and reweighting the corresponding text embeddings to maintain consistent identity features across frames. 
A more fundamental class of training-free methods promotes consistency by reusing character-corresponding key-value (KV) representations within the attention layers of diffusion models \cite{consistory, zhou2024storydiffusion, he2024dreamstory, characonsist}. By enabling different images to attend to shared KV features associated with the same character, these methods directly enforce appearance consistency at the feature level.
However, these methods typically require prompts to follow a strict \textit{character + activity} format, and batch generation further increases memory consumption during inference.
CharaConsist~\cite{characonsist} proposes to use the more powerful Flux \cite{flux} model and supports both foreground-consistent and background-consistent image synthesis. These design choices impose stricter prompt requirements, as the method requires explicit background descriptions and is limited to single-character scenarios.

In contrast, our method enables training-free visual storytelling under free-form prompts.
Instead of relying on repeated description or structured prompt formats, we introduce an entity grounding module to automatically associate reference mentions with their corresponding character descriptions.
Furthermore, our formulation naturally supports multi-character stories, where multiple characters may appear and interact across frames, while existing KV reuse methods are primarily designed for structured prompts and single-character scenarios.

\section{Method}

\begin{figure}[tb]
    \centering
    \includegraphics[width=\linewidth]{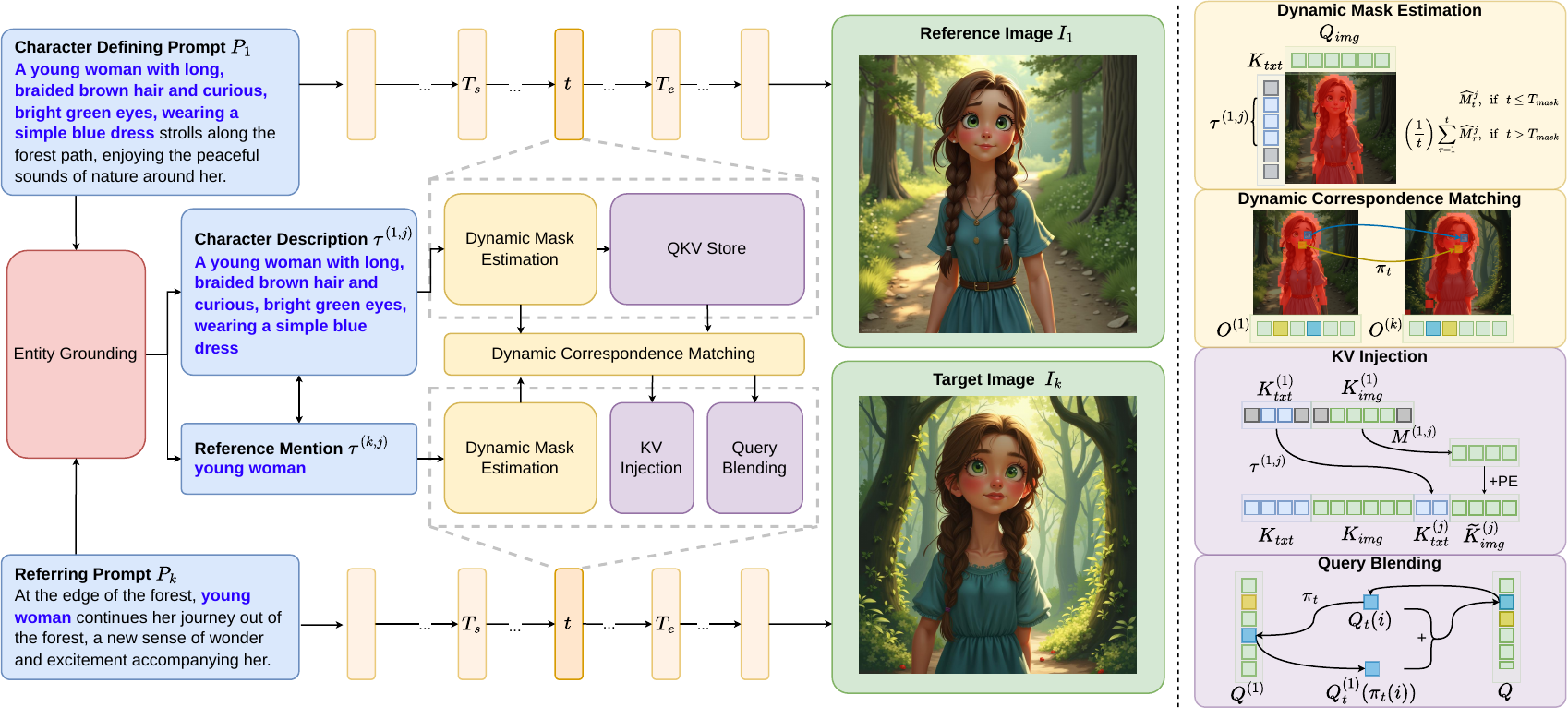}
\caption{
\textbf{Overview of our proposed FreeStory framework.}
Given a character-defining prompt $P_1$, the model generates the reference image $I_1$ and uses entity grounding to associate the character description $\tau^{(1,j)}$ with reference mentions $\tau^{(k,j)}$ in referring prompt $P_k$ for character $c^{(j)}$.
During generation of $I_1$, we extract cross-attention weights to compute the dynamic mask $\tilde{M}_t^{(1,j)}$ and store the corresponding key, value, and query features in a QKV bank.
Given a referring prompt $P_k$, the model generates the target image $I_k$ and extracts its character mask.
Token-level correspondence between $I_1$ and $I_k$ is established within the masks, enabling correspondence-guided key-value injection with positional re-encoding (+PE) and query blending.
This process propagates character identity from $I_1$ to $I_k$, ensuring consistent appearance under free-form prompts.
}
    \label{fig:overview}
\end{figure}
Our method enables character-consistent visual storytelling under both structured and free-form prompt settings in a fully training-free manner.
\Cref{fig:overview} provides an overview of our pipeline.
Given a sequence of prompts, we first identify the character description span in the initial prompt and generate the first image, referred to as the \emph{reference image}, while extracting character-aware attention masks and storing the corresponding attention features.
For each subsequent prompt, we locate the character reference and generate the image, referred to as a \emph{target image}, while establishing correspondence between the target image and the reference image based on attention similarity.
We then reuse the stored attention features via correspondence-guided KV injection and query blending within a specific timestep window, allowing the model to attend to consistent character information across images.
This process preserves character identity while maintaining flexibility for pose, expression, and scene variations.


\subsection{Problem Setup}

Given a story consisting of a sequence of prompts $\{P_1, P_2, \dots, P_N\}$, the goal of visual storytelling is to generate a sequence of images $\{I_1, I_2, \dots, I_N\}$, where each image $I_i$ corresponds to prompt $P_i$ and accurately reflects its content.
A key requirement is to maintain character consistency across all generated images.

Each story contains one or more characters.
In the first prompt $P_1$, each character is introduced through a descriptive text span that defines its appearance, referred to as the \emph{character description}.
In the structured storytelling setting used in previous work, this character description is required to appear at the beginning of every prompt, explicitly repeating the full description in each image.
In contrast, we consider a more realistic \emph{free-form storytelling} setting.
We assume that each character description appears only once in the first prompt $P_1$, without any restriction on where it is in the prompt.
In subsequent prompts $\{P_2, \dots, P_N\}$, characters are implicitly referred to using pronouns or type-based expressions instead of repeating the full description.

Furthermore, in the multi-character setting, not all characters are required to appear in every subsequent prompt, reflecting realistic story progression.
Under this setup, the challenge is to maintain consistent character identity across images despite the absence of explicit character descriptions in later prompts.

\subsection{Cross-image Character Correspondence}

To enable consistent character generation, we establish correspondence between character regions of the reference image and each target image during the diffusion process.
This correspondence is dynamically computed within a timestep range $[T_s, T_e]$ based on attention features and character masks.

\subsubsection{Character-aware Attention Mask Extraction}

Given a prompt $P_i$, we extract character masks on latent tokens during the generation of the image $I_i$.
This procedure is applied uniformly to both the reference image and all target images.

At diffusion timestep $t$ and attention layer $l$, the model computes the cross-attention between latent tokens and text tokens:
\begin{equation}
A_{t,l} =
\mathrm{Softmax}
\left(
{Q_{t,l,img} K_{t,l,txt}^\top} / {\sqrt{d}}
\right),
\end{equation}
Each character $c^{(j)}$, where $c^{(j)}$ denotes the $j$-th character, is associated with a set of text token indices $\mathcal{T}^{(j)}$ corresponding to its description in the prompt.
We aggregate cross-attention weights over character tokens and average across all $L$ layers to obtain the soft character mask, which is then converted into a binary mask using Otsu thresholding \cite{otsu}:
\begin{equation}
\hat{M}_t^{(j)}
=
\frac{1}{L}
\sum_{l=1}^{L}
\sum_{x \in \mathcal{T}^{(j)}}
A_{t,l}[:, x], 
\qquad
M_t^{(j)}
=
\mathrm{Otsu}
\left(
\hat{M}_t^{(j)}
\right).
\end{equation}

\paragraph{Dynamic Attention Mask Estimation. }

Previous works \cite{consistory,characonsist} typically perform a separate pre-generation pass to extract attention masks at a fixed timestep or by averaging attention weights across multiple timesteps.
The extracted mask is then reused in a second full generation process.
This approach increases inference cost and assumes that a fixed mask is optimal for all timesteps.

To improve mask stability across timesteps, we adopt a dynamic mask estimation strategy.
Empirically, attention responses exhibit different behaviors at early and late diffusion steps.
Based on this observation, we compute the character mask using the attention map from the current timestep for early steps, and use the average attention map over previous timesteps for later steps.
Detailed analysis and quantitative evaluation of mask quality across timesteps are provided in the Appendix.
The soft mask before the final Otsu thresholding is defined as the following piecewise function:
\begin{equation}
\tilde{M}_t^{(j)} =
\begin{cases}
\hat{M}_t^{(j)}, & t \leq T_{mask}, \\[6pt]
\frac{1}{t}
\sum_{\tau = 1}^{t}
\hat{M}_\tau^{(j)}, & t > T_{mask}.
\end{cases}
\end{equation}

This dynamic strategy enables accurate mask extraction throughout the diffusion process using a single generation pass, eliminating the need for costly pre-generation while improving mask quality at each timestep.
The extracted masks not only localize character regions, but also provide the spatial support for correspondence matching and feature reuse in subsequent stages.

\subsubsection{Attention Correspondence Matching. }

The goal of correspondence matching is to establish spatial interaction between the target and the reference image at the character level, which serves as the foundation for subsequent feature reuse, including KV injection and query blending.

Establishing correspondence between image regions using diffusion features has been explored in prior work \cite{dift,ditft}.
Several visual storytelling methods also use similar strategies to support feature injection across images \cite{consistory,characonsist}.
The core idea is to identify corresponding latent tokens within the character regions based on feature similarity.
As a result, correspondence quality critically depends on the accuracy of the character masks, which motivates our dynamic mask extraction strategy described in the previous subsection.

Given the attention output features $O_{t,l}$ at timestep $t$ and layer $l$, we first average the attention output across all layers and get  $\bar{O}_t^{(1)}$ and $\bar{O}_t^{(k)}$ for the reference and target images, respectively: $
\bar{O}_t =
\frac{1}{L}
\sum_{l=1}^{L}
O_{t,l},
$
where $L$ is the number of cross-attention layers. We then compute pairwise cosine similarity between latent tokens within the character mask regions:

\begin{equation}
S^{(j)}_{pq}
=
\cos
\left(
\bar{O}_t^{(k)}(p),
\bar{O}_t^{(1)}(q)
\right),
\quad
p \in \mathcal{M}_t^{(k,j)},\;
q \in \mathcal{M}_t^{(1,j)}.
\end{equation}

The similarity computation is restricted to latent tokens within the character mask regions.
For each latent token $p$ in the target image mask, we identify the most similar token in the reference image:
\begin{equation}
\pi_t(p) =
\arg\max_{p \in \mathcal{M}^{(k,j)}_t}
S_{pq},
\quad q \in \mathcal{M}^{(1,j)}_t,
\end{equation}
where $\mathcal{M}^{(1,j)}_t$ and $\mathcal{M}^{(k,j)}_t$ are the mask of the reference and target image for character $c^{(j)}$, respectively.
This correspondence function $\pi_t$ establishes token-level interaction between the character regions of the two images.

\paragraph{Dynamic Correspondence Matching.}

Since character masks are dynamically extracted and evolve during the diffusion process, the correspondence relationship should also adapt accordingly.
For early timesteps ($t \leq T_{corr}$), we recompute correspondence at every timestep, allowing the model to progressively refine the correspondence as mask quality improves.
For later timesteps ($t > T_{corr}$), instead of recomputing correspondence from scratch, we update them based on the evolving mask region while preserving the previously established mappings.

The dynamic mask extraction and correspondence matching together establish and refine the spatial interaction between the current frame and the reference frame throughout the generation process.
This interaction provides precise token-level alignment between character regions, enabling effective propagation of character appearance in the feature reuse stage.

\subsection{Attention-based Feature Reuse}

Given the correspondence between the reference image and the target image established in the previous subsection, we reuse character-specific attention features to guide generation and preserve character identity.
These operations are performed within a timestep window $[T_s, T_e]$, where semantic structure is sufficiently formed while still allowing controllable generation.

\subsubsection{Key-Value Injection}

Our method is built upon the single-stream attention architecture in FLUX, where text tokens and image latent tokens are concatenated into a unified sequence and processed by self-attention.
This design allows direct interaction between textual and visual features within a single attention layer.
Motivated by this property, we reuse character-specific features from the reference image in \emph{both} domains: (i) the image tokens within the character mask, and (ii) the text tokens corresponding to the character description span.

At diffusion timestep $t$, for each character $c^{(j)}$, we extract reference features from the generation of the reference image $I_1$:
\begin{equation}
K^{(j)}_{t,\mathrm{img}} = K^{(1)}_t(\mathcal{M}^{(1,j)}_t), \qquad
V^{(j)}_{t,\mathrm{img}} = V^{(1)}_t(\mathcal{M}^{(1,j)}_t),
\end{equation}
\begin{equation}
K^{(j)}_{t,\mathrm{txt}} = K^{(1)}_t(\mathcal{T}^{(1,j)}), \qquad
V^{(j)}_{t,\mathrm{txt}} = V^{(1)}_t(\mathcal{T}^{(1,j)}),
\end{equation}
where $\mathcal{M}^{(1,j)}_t$ denotes the latent-token character mask of $c^{(j)}$ in the reference image, and $\mathcal{T}^{(1,j)}$ denotes the character description span for $c^{(j)}$.

Since FLUX relies on positional information for spatial reasoning, we re-encode the positional embeddings of the injected \emph{image} key features using the correspondence function $\pi_t^{(j)}$ between the target image and the reference image:
\begin{equation}
\tilde{K}^{(j)}_{t,\mathrm{img}} = \mathrm{PE}\!\left(K^{(j)}_{t,\mathrm{img}}, \pi_t^{(j)}\right),
\end{equation}
where $\mathrm{PE}(\cdot)$ denotes positional encoding of keys according to the matched reference.
For the \emph{text} features, we directly reuse the keys and values.
We inject all characters' reference features by concatenating them to the target features:
\begin{equation}
\tilde{K}_t =
\mathrm{Concat}
\Big(
K_t,\;
K^{(1)}_{t,\mathrm{txt}},\tilde{K}^{(1)}_{t,\mathrm{img}},\;
\dots,\;
K^{(J)}_{t,\mathrm{txt}},\tilde{K}^{(J)}_{t,\mathrm{img}}
\Big),
\end{equation}
\begin{equation}
\tilde{V}_t =
\mathrm{Concat}
\Big(
V_t,\;
{V}^{(1)}_{t,\mathrm{txt}},{V}^{(1)}_{t,\mathrm{img}},\;
\dots,\;
{V}^{(J)}_{t,\mathrm{txt}},{V}^{(J)}_{t,\mathrm{img}}
\Big),
\end{equation}
where $J$ is the number of characters in both $P_1$ and current prompt.

To ensure correct feature interaction, we construct a modified attention mask that restricts the injected features to be visible only to the corresponding character tokens in both image and text domains.
Concretely, for each character $c_j$, only the target tokens belonging to $c_j$ (i.e., in $\mathcal{M}^{(k,j)}_t$ and in the character reference span of $P_k$) are allowed to attend to the injected tokens $\{K^{(j)}_{t,\mathrm{txt}}, \tilde{K}^{(j)}_{t,\mathrm{img}}\}$, while attention to injected features from other characters is masked out.
This prevents interference with unrelated regions and maintains semantic consistency.

\subsubsection{Query Blending}

Prior work often injects attention output features directly to enforce consistency.
However, this introduces a strong constraint on the generated features, often resulting in overly similar appearances across images and reduced diversity. 
Instead, we modify the query features based on correspondence.
Since the query determines how attention is computed over key and value features, blending query features allows the model to align attention patterns while preserving flexibility in the attended content. Formally, we blend the query features using the correspondence function:

\begin{equation}
\tilde{Q}_t(i) =
\lambda Q_t(i)
+
(1-\lambda) Q_t^{(1)}(\pi_t(i)),
\end{equation}
where $\lambda$ controls the balance between preserving current generation and reusing reference features. This blending is applied only within the character mask region. Finally, the modified attention is computed as:

\begin{equation}
\tilde{A}_t =
\mathrm{Softmax}
\left(
\tilde{Q}_t \tilde{K}_t^{\top} / \sqrt{d}
\right), \quad
\tilde{O}_t =
\tilde{A}_t \tilde{V}_t.
\end{equation}

By modifying the attention computation in this manner, our method propagates character identity from the reference image to the target image while maintaining natural variations.

\subsection{Entity Grounding Across Prompts}

Prior visual storytelling datasets use a structured prompt format, where the full character description is explicitly repeated in every prompt.
It deviates from natural storytelling, where characters are typically introduced once and subsequently referred to using pronouns or type-based expressions.
To better reflect realistic narrative settings, we consider a free-form storytelling setup, in which character descriptions appear only in the first prompt, and subsequent prompts refer to characters implicitly.

This setup introduces a new challenge: the model must identify which character a reference mention corresponds to, without relying on explicit repetition of the character description.
Formally, given the character-defining prompt $P_1$ and a referring prompt $P_k$, the goal of entity grounding is to identify all reference mentions in $P_k$ and associate each reference mention with its corresponding character description token indices in $P_1$.
Let $\mathcal{T}^{(k,j)}$ denote the text token indices of a reference mention of character $c^{(j)}$ in prompt $P_k$, and let $\mathcal{T}^{(1,j)}$ denote the description of character $c^{(j)}$ in the character-defining prompt $P_1$.
The entity grounding task is to establish a mapping: $\mathcal{T}^{(k,j)} \;\rightarrow\; \mathcal{T}^{(1,j)}$, which enables consistent character identification across prompts.

To perform this grounding, we use an off-the-shelf linguistic coreference resolution system provided by Stanza~\cite{stanza}.
Stanza is a neural pipeline for natural language processing that includes coreference resolution, which identifies textual expressions referring to the same entity across sentences.
Given the concatenated story prompts, Stanza produces coreference chains that link reference mentions to their representative mentions.
Using these coreference chains, we build up the mapping from $\mathcal{T}^{(k,j)}$ to $\mathcal{T}^{(1,j)}$.
These recovered token indices are then used for character mask extraction and subsequent attention-based feature reuse.

Importantly, this grounding process operates fully automatically and does not rely on any ground-truth annotations.
This enables our method to generalize to free-form storytelling settings.

\subsection{Dataset}
\label{sec:dataset}
In addition to the aforementioned rigid prompt format, existing benchmarks also lack explicit visual grounding, making it difficult to isolate a model's linguistic comprehension from its generative capabilities. To rigorously evaluate a model's capabilities in maintaining character consistency and resolving complex co-references over long contexts, we introduce \textit{FreeStoryBench}. We constructed a structured corpus of 200 unique 6-scene narrative storylines, which can be procedurally expanded into 1,200 distinct textual variations. We employ a novel ``Dual-Generation'' pipeline that decouples the underlying semantic logic of a story (e.g., entity tracking via explicit IDs) from its textual surface realization. This allows us to map a single ground-truth narrative into six progressive difficulty tiers; ranging from free-form format where characters are referenced with pronouns or type-based mentions to redundant visual descriptions. By controlling the linguistic complexity while keeping the visual requirements constant, our proposed dataset enables precise, stratified evaluation of generative models. Comprehensive details regarding the generation pipeline, prompt configurations, and full dataset examples are provided in  the Appendix. 

\section{Experiments}

\begin{table}[tb]
\caption{
\textbf{Quantitative results on the structured ConsiStory+ dataset.}
Our method achieves the best character consistency across CLIP-I, DINO, and DreamSim among training-free storytelling methods, demonstrating the effectiveness of our FreeStory. Flux.1-Kontext also achieves strong consistency due to reference-based image-editing. Bold means the best.
}    \label{tab:main}
    \centering
    \begin{tabular}{l|r|r|r|r}
    \toprule
    \textbf{Method} & \multicolumn{1}{l|}{\textbf{CLIP-T} $\uparrow$} & \multicolumn{1}{l|}{\textbf{CLIP-I} $\uparrow$} & \multicolumn{1}{l|}{\textbf{DINO} $\uparrow$} & \multicolumn{1}{l}{\textbf{DreamSim} $\downarrow$} \\
    \midrule
    SDXL           & 0.8584          & 0.8604          & 0.5959          & 0.3331          \\
    Flux.1         & 0.8359          & 0.8784          & 0.6487          & 0.2703          \\
    Flux.1-Kontext & 0.8179          & 0.9326          & 0.8325          & 0.1371          \\
    \midrule
    ConsiStory     & \textbf{0.8564} & 0.8906          & 0.6800          & 0.2479          \\
    StoryDiffusion & 0.8452          & 0.8916          & 0.6959          & 0.2626          \\
    1P1S           & 0.8193          & 0.8950          & 0.6955          & 0.2345          \\
    FreeStory (Ours) & 0.8369          & \textbf{0.8979} & \textbf{0.7118} & \textbf{0.2234} \\
    \bottomrule
    \end{tabular}
\end{table}

\begin{figure}[!t]
    \centering
    \includegraphics[width=\linewidth]{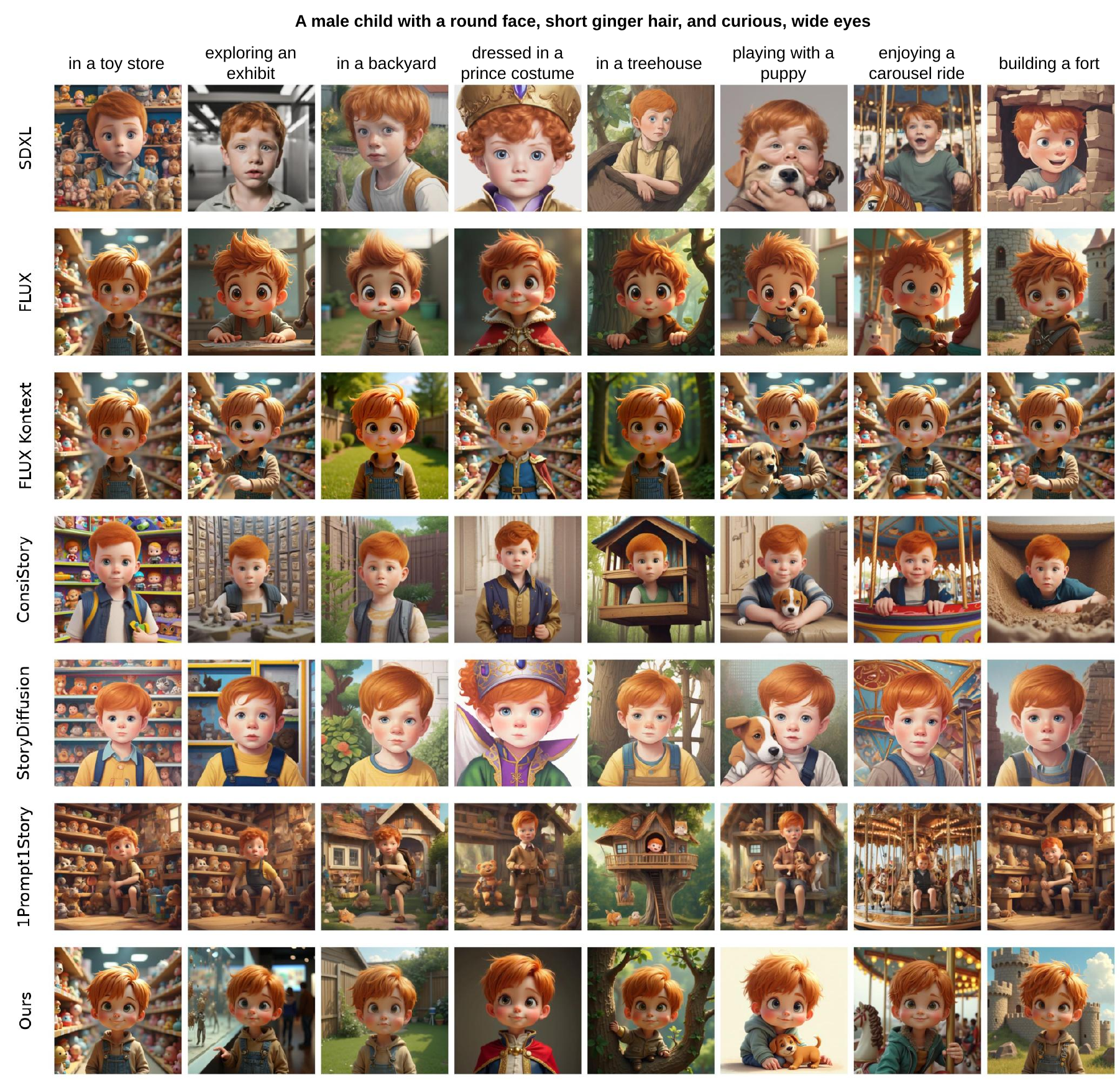}
    \caption{\textbf{Qualitative results on ConsiStory+ dataset.} Independent generation using SDXL and FLUX.1 fails to preserve character identity. Prior storytelling methods partially improve consistency but still suffer from identity drift. FLUX.1-Kontext achieves strong appearance similarity but exhibits copy-paste artifacts, with limited pose and background diversity. Our method preserves character identity while maintaining natural motion and scene variation, resulting in more coherent and realistic storytelling.}
    \label{fig:120_humans_child}
\end{figure}

\subsection{Experimental Setup}

\paragraph{Dataset.}
We conduct experiments on two datasets, \textit{ConsiStory+} \cite{onepromptonestory} and our proposed \textit{FreeStoryBench}. 
\textit{ConsiStory+} contains 192 single-character stories and 1100 images in total, where each story follows the structured prompt format with explicit character descriptions. 
\textit{FreeStoryBench} contains 100 single-character and 100 multi-character stories, with 6 images per story. 

\paragraph{Metrics.}
We evaluate text-image alignment using CLIP-T \cite{clip} similarity.
To assess character consistency, we use three image-image similarity metrics, including CLIP-I \cite{clip}, DINO \cite{dino}, and DreamSim \cite{dreamsim}.
For all consistency metrics, we first use Grounded SAM \cite{ren2024grounded} to segment the character region from the image, and replace the background with random noise to eliminate background influence.
The final score for each story is computed as the average similarity across all frame pairs within the story.
Compared to prior work that uses CarveKit for background removal, we find Grounded SAM provides more accurate and robust character segmentation. Detailed analysis is provided in the Appendix.

\paragraph{Baselines.}
We compare our method with two pretrained text-to-image diffusion models, SDXL \cite{sdxl} and FLUX.1 \cite{flux}, where images are generated independently for each prompt.
We also include FLUX.1-Kontext \cite{flux1kontext}, an image editing model designed for strong character preservation.
Specifically, we first generate the first image using FLUX.1, and use it as the reference image for the subsequent images.
In addition, we compare with three visual storytelling baselines, including ConsiStory \cite{consistory}, StoryDiffusion \cite{zhou2024storydiffusion}, and 1Prompt1Story \cite{onepromptonestory}.


\subsection{Results on Structured Storytelling Benchmark}

\Cref{tab:main} presents the quantitative results on the ConsiStory+ dataset under the structured prompt setting.
Our method achieves the best performance across all three character consistency metrics, including CLIP-I, DINO, and DreamSim, outperforming all prior training-free storytelling baselines, namely ConsiStory, StoryDiffusion, and 1Prompt1Story.
This demonstrates the effectiveness of our attention-based feature reuse mechanism in preserving character identity across frames without requiring additional training.
Compared to the base text-to-image models, SDXL and FLUX.1, which generate each frame independently, our method significantly improves character consistency, while maintaining competitive text-image alignment.
This confirms that the consistency improvements arise from our proposed method rather than the underlying diffusion model.

We also compare with FLUX-Kontext, a training-based image editing model designed for strong character preservation.
While FLUX-Kontext achieves higher consistency scores, we observe that it often suffers from a copy-paste effect, where characters exhibit highly similar poses and appearances across frames, and backgrounds remain nearly identical.
This leads to reduced motion and scene diversity, as shown in the qualitative examples~\cref{fig:120_humans_child}.
Such artifacts are not fully reflected by similarity-based metrics, which primarily measure appearance similarity but do not capture diversity or naturalness.

In contrast, our method preserves character identity while allowing natural variations in pose, expression, and background, resulting in more realistic storytelling sequences.
Qualitative comparisons~\cref{fig:120_humans_child} further demonstrate that our method produces substantially more consistent characters than prior storytelling baselines, while avoiding the copy-paste artifacts observed in FLUX-Kontext. 
More qualitative comparison can be found in the Appendix.

\begin{table}[tb]
\centering
\caption{
\textbf{Quantitative results on FreeStoryBench under type-based referring prompts.}
GT Grounding uses ground-truth correspondence, while Entity Grounding uses our automatic entity grounding module. Best and second-best results are shown in \textbf{bold} and \underline{underlined}, respectively.
}\label{tab:freestorybench}
\begin{tabular}{c|l|c|c|c|c}
\toprule
 & \textbf{Method} & \textbf{CLIP-T} $\uparrow$ & \textbf{CLIP-I} $\uparrow$ & \textbf{DINO} $\uparrow$ & \textbf{DreamSim} $\downarrow$ \\
\midrule
\multirow{6}{*}{\rotatebox[origin=c]{90}{\begin{tabular}{c}
Single \\
Character
\end{tabular}}}
& FLUX & 0.8193 & 0.8242 & 0.5123 & 0.3825 \\
& ConsiStory (GT Grounding) & 0.8457 & 0.8521 & 0.5918 & 0.3239 \\
& StoryDiffusion (GT Grounding) & 0.8184 & 0.8560 & 0.5950 & 0.3651 \\
& 1P1S (GT Grounding) & 0.8218 & 0.8467 & 0.5655 & 0.3683 \\
& Ours (GT Grounding) & \textbf{0.8550} & \underline{0.8618} & \textbf{0.6243} & \textbf{0.2850} \\
& Ours (Entity Grounding) & \underline{0.8477} & \textbf{0.8638} & \underline{0.6210} & \underline{0.2886} \\
\midrule
\multirow{6}{*}{\rotatebox[origin=c]{90}{\begin{tabular}{c}
Multi \\
Character
\end{tabular}}}
& FLUX & 0.8291 & 0.8271 & 0.4445 & 0.4177 \\
& ConsiStory (GT Grounding) & \underline{0.8599} & 0.8330 & 0.4789 & 0.3833 \\
& StoryDiffusion (GT Grounding) & 0.8374 & \textbf{0.8467} & 0.4909 & 0.4067 \\
& 1P1S (GT Grounding) & 0.8281 & 0.8345 & 0.4365 & 0.4108 \\
& Ours (GT Grounding) & \textbf{0.8604} & 0.8435 & \textbf{0.5439} & \textbf{0.3482} \\
& Ours (Entity Grounding) & 0.8564 & \underline{0.8440} & \underline{0.5397} & \underline{0.3504} \\
\bottomrule
\end{tabular}
\end{table}

\subsection{Results on Free-form Storytelling Benchmark}

\Cref{tab:freestorybench} presents quantitative results on our proposed FreeStoryBench dataset, where the character-defining prompt contains the full description span and all referring prompts use only type-based reference mentions.
To enable attention-based feature reuse, the correspondence between the reference mention in each referring prompt and the description in the character-defining prompt needs to be identified.
Our dataset provides these correspondences as annotations for evaluation, but in realistic settings they are not available and must be recovered automatically.
To evaluate this aspect, we consider two variants of our method.
\textbf{GT Grounding} uses ground-truth annotations provided by the dataset.
This setting serves as an oracle reference and is primarily included to enable a fair comparison with the ConsiStory, StoryDiffusion and 1P1S, which require explicit correspondence information to operate.
In contrast, \textbf{Entity Grounding} uses our automatic grounding module based on Stanza to recover this correspondence without manual supervision.

For single-character stories, our method significantly improves character consistency over the base diffusion model FLUX across all metrics, demonstrating the effectiveness of attention-based feature reuse under free-form prompts.
Our method also consistently outperforms all baselines, including ConsiStory \cite{consistory}, StoryDiffusion \cite{zhou2024storydiffusion}, and 1Prompt1Story (1P1S) \cite{onepromptonestory}.
Importantly, the performance of our method using Entity Grounding is comparable to the oracle GT Grounding variant, with only marginal differences across metrics.
This result shows that our entity grounding module can reliably recover character correspondence automatically, enabling fully automatic free-form visual storytelling without requiring ground-truth annotations.
In contrast, the baseline methods relie on explicit character-description correspondence and therefore cannot operate in realistic free-form settings without such annotations.

We further evaluate our method on the multi-character subset of FreeStoryBench and observe consistent conclusions in \cref{tab:freestorybench}.
Both the GT Grounding and Entity Grounding variants outperform the base FLUX model and achieve stronger overall performance than the compared baselines.
These results demonstrate that our method effectively extends to more complex multi-character scenarios.
\Cref{fig:demo} shows a multi-character example and additional qualitative results are provided in the Appendix.

We note that the overall performance in the multi-character setting is slightly lower than in the single-character case.
This is expected, as multi-character storytelling introduces additional challenges in both entity grounding and attention mask extraction.
Specifically, the model must correctly associate multiple reference mentions with their corresponding description spans, while simultaneously extracting accurate and disentangled character masks for each character during generation.

Because the compared methods use different diffusion backbones, we report the relative inference overhead of each method with respect to its corresponding backbone.
Over SDXL, 1P1S, StoryDiffusion, and ConsiStory require $1.09\times$, $3.09\times$, and $2.24\times$ the backbone memory, respectively, and $1.94\times$, $1.38\times$, and $3.47\times$ the backbone inference time.
In comparison, our method requires $1.14\times$ the memory and $1.51\times$ the inference time of FLUX, introducing only moderate computational overhead relative to its backbone.

Overall, our method consistently outperforms the base model and all compared storytelling baselines in both single- and multi-character settings, while only introducing moderate computational overhead.

\section{Conclusion}
In this work, we introduced free-form visual storytelling, a more realistic setting in which characters are described once and subsequently referred to through implicit or abbreviated mentions.
We proposed FreeStory, a training-free framework that formulates character consistency under free-form prompts as entity-grounded feature reuse, connecting linguistic reference resolution with attention-based visual feature transfer.
We further introduced FreeStoryBench, a benchmark supporting both single- and multi-character narratives under free-form prompt conditions.
Extensive experiments demonstrate that FreeStory achieves state-of-the-art performance among training-free methods on structured benchmarks and stronger overall consistency than the baselines under free-form prompts.

\clearpage
\bibliographystyle{plainnat}
\bibliography{main}

@INPROCEEDINGS{arldm,
  author={Pan, Xichen and Qin, Pengda and Li, Yuhong and Xue, Hui and Chen, Wenhu},
  booktitle={2024 IEEE/CVF Winter Conference on Applications of Computer Vision (WACV)}, 
  title={Synthesizing Coherent Story with Auto-Regressive Latent Diffusion Models}, 
  year={2024},
  volume={},
  number={},
  pages={2908-2918},
  doi={10.1109/WACV57701.2024.00290}}

@INPROCEEDINGS{makeastory,
  author={Rahman, Tanzila and Lee, Hsin- Ying and Ren, Jian and Tulyakov, Sergey and Mahajan, Shweta and Sigal, Leonid},
  booktitle={2023 IEEE/CVF Conference on Computer Vision and Pattern Recognition (CVPR)}, 
  title={Make-A-Story: Visual Memory Conditioned Consistent Story Generation}, 
  year={2023},
  volume={},
  number={},
  pages={2493-2502},
  doi={10.1109/CVPR52729.2023.00246}}

@inproceedings{talecrafter,
author = {Gong, Yuan and Pang, Youxin and Cun, Xiaodong and Xia, Menghan and He, Yingqing and Chen, Haoxin and Wang, Longyue and Zhang, Yong and Wang, Xintao and Shan, Ying and Yang, Yujiu},
title = {Interactive Story Visualization with Multiple Characters},
year = {2023},
isbn = {9798400703157},
publisher = {Association for Computing Machinery},
address = {New York, NY, USA},
doi = {10.1145/3610548.3618184},
booktitle = {SIGGRAPH Asia 2023 Conference Papers},
articleno = {101},
numpages = {10},
location = {Sydney, NSW, Australia},
series = {SA '23}
}

@inproceedings{acmvsg,
    title = "Improved Visual Story Generation with Adaptive Context Modeling",
    author = "Feng, Zhangyin  and
      Ren, Yuchen  and
      Yu, Xinmiao  and
      Feng, Xiaocheng  and
      Tang, Duyu  and
      Shi, Shuming  and
      Qin, Bing",
    editor = "Rogers, Anna  and
      Boyd-Graber, Jordan  and
      Okazaki, Naoaki",
    booktitle = "Findings of the Association for Computational Linguistics: ACL 2023",
    month = jul,
    year = "2023",
    address = "Toronto, Canada",
    publisher = "Association for Computational Linguistics",
    url = "https://aclanthology.org/2023.findings-acl.305",
    pages = "4939--4955",
}

@InProceedings{storygen,
    author    = {Liu, Chang and Wu, Haoning and Zhong, Yujie and Zhang, Xiaoyun and Wang, Yanfeng and Xie, Weidi},
    title     = {Intelligent Grimm - Open-ended Visual Storytelling via Latent Diffusion Models},
    booktitle = {Proceedings of the IEEE/CVF Conference on Computer Vision and Pattern Recognition (CVPR)},
    month     = {June},
    year      = {2024},
    pages     = {6190-6200}
}

@article{cogcartoon,
author = {Zhu, Zhongyang and Tang, Jie},
title = {CogCartoon: Towards Practical Story Visualization},
year = {2024},
issue_date = {Apr 2025},
publisher = {Kluwer Academic Publishers},
address = {USA},
volume = {133},
number = {4},
issn = {0920-5691},
doi = {10.1007/s11263-024-02267-5},
journal = {Int. J. Comput. Vision},
month = oct,
pages = {1808–1833},
numpages = {26}
}

@INPROCEEDINGS{causalstory,
  author={Song, Tianyi and Cao, Jiuxin and Wang, Kun and Liu, Bo and Zhang, Xiaofeng},
  booktitle={ICASSP 2024 - 2024 IEEE International Conference on Acoustics, Speech and Signal Processing (ICASSP)}, 
  title={Causal-Story: Local Causal Attention Utilizing Parameter-Efficient Tuning for Visual Story Synthesis}, 
  year={2024},
  volume={},
  number={},
  pages={3350-3354},
  doi={10.1109/ICASSP48485.2024.10446420}}

@inproceedings{storyimager,
author = {Tao, Ming and Bao, Bing-Kun and Tang, Hao and Wang, Yaowei and Xu, Changsheng},
title = {StoryImager: A Unified and Efficient Framework for Coherent Story Visualization and Completion},
year = {2024},
isbn = {978-3-031-72991-1},
publisher = {Springer-Verlag},
address = {Berlin, Heidelberg},
url = {https://doi.org/10.1007/978-3-031-72992-8_27},
booktitle = {Computer Vision – ECCV 2024: 18th European Conference, Milan, Italy, September 29–October 4, 2024, Proceedings, Part LVI},
pages = {479–495},
numpages = {17},
location = {Milan, Italy}
}

@article{yang2024seedstory,
  title={SEED-Story: Multimodal Long Story Generation with Large Language Model}, 
  author={Shuai Yang and Yuying Ge and Yang Li and Yukang Chen and Yixiao Ge and Ying Shan and Yingcong Chen},
  year={2024},
  journal={arXiv preprint arXiv:2407.08683},
  url={https://arxiv.org/abs/2407.08683}, 
}

@INPROCEEDINGS{storygptv,
  author={Shen, Xiaoqian and Elhoseiny, Mohamed},
  booktitle={2025 IEEE/CVF Conference on Computer Vision and Pattern Recognition (CVPR)}, 
  title={StoryGPT-V: Large Language Models as Consistent Story Visualizers}, 
  year={2025},
  volume={},
  number={},
  pages={13273-13283},
  doi={10.1109/CVPR52734.2025.01239}}

@inproceedings{
zhou2024storydiffusion,
title={StoryDiffusion: Consistent Self-Attention for Long-Range Image and Video Generation},
author={Yupeng Zhou and Daquan Zhou and Ming-Ming Cheng and Jiashi Feng and Qibin Hou},
booktitle={The Thirty-eighth Annual Conference on Neural Information Processing Systems},
year={2024},
url={https://openreview.net/forum?id=VFqzxhINFU}
}

@inproceedings{thechosenone,
author = {Avrahami, Omri and Hertz, Amir and Vinker, Yael and Arar, Moab and Fruchter, Shlomi and Fried, Ohad and Cohen-Or, Daniel and Lischinski, Dani},
title = {The Chosen One: Consistent Characters in Text-to-Image Diffusion Models},
year = {2024},
isbn = {9798400705250},
publisher = {Association for Computing Machinery},
address = {New York, NY, USA},
url = {https://doi.org/10.1145/3641519.3657430},
booktitle = {ACM SIGGRAPH 2024 Conference Papers},
articleno = {26},
numpages = {12},
location = {Denver, CO, USA},
series = {SIGGRAPH '24}
}

@article{zheng2024temporalstory,
  title={TemporalStory: Enhancing Consistency in Story Visualization using Spatial-Temporal Attention},
  author={Zheng, Sixiao and Fu, Yanwei},
  journal={arXiv preprint arXiv:2407.09774},
  year={2024}
}

@article{he2024dreamstory,
  title={DreamStory: Open-Domain Story Visualization by LLM-Guided Multi-Subject Consistent Diffusion},
  author={He, Huiguo and Yang, Huan and Tuo, Zixi and Zhou, Yuan and Wang, Qiuyue and Zhang, Yuhang and Liu, Zeyu and Huang, Wenhao and Chao, Hongyang and Yin, Jian},
  journal={arXiv preprint arXiv:2407.12899},
  year={2024}
}

@article{wang2024oneactor,
  title={OneActor: Consistent Character Generation via Cluster-Conditioned Guidance},
  author={Wang, Jiahao and Yan, Caixia and Lin, Haonan and Zhang, Weizhan},
  journal={arXiv preprint arXiv:2404.10267},
  year={2024}
}

@article{consistory,
author = {Tewel, Yoad and Kaduri, Omri and Gal, Rinon and Kasten, Yoni and Wolf, Lior and Chechik, Gal and Atzmon, Yuval},
title = {Training-Free Consistent Text-to-Image Generation},
year = {2024},
issue_date = {July 2024},
publisher = {Association for Computing Machinery},
address = {New York, NY, USA},
volume = {43},
number = {4},
issn = {0730-0301},
doi = {10.1145/3658157},
journal = {ACM Trans. Graph.},
month = jul,
articleno = {52},
numpages = {18}
}

@inproceedings{onepromptonestory,
title={One-Prompt-One-Story: Free-Lunch Consistent Text-to-Image Generation Using a Single Prompt},
author={Tao Liu and Kai Wang and Senmao Li and Joost van de Weijer and Fahad Shahbaz Khan and Shiqi Yang and Yaxing Wang and Jian Yang and Ming-Ming Cheng},
booktitle={The Thirteenth International Conference on Learning Representations},
year={2025},
url={https://openreview.net/forum?id=cD1kl2QKv1}
}

@INPROCEEDINGS{sd,
  author={Rombach, Robin and Blattmann, Andreas and Lorenz, Dominik and Esser, Patrick and Ommer, Björn},
  booktitle={2022 IEEE/CVF Conference on Computer Vision and Pattern Recognition (CVPR)}, 
  title={High-Resolution Image Synthesis with Latent Diffusion Models}, 
  year={2022},
  volume={},
  number={},
  pages={10674-10685},
  doi={10.1109/CVPR52688.2022.01042}}

@inproceedings{sdxl,
title={{SDXL}: Improving Latent Diffusion Models for High-Resolution Image Synthesis},
author={Dustin Podell and Zion English and Kyle Lacey and Andreas Blattmann and Tim Dockhorn and Jonas M{\"u}ller and Joe Penna and Robin Rombach},
booktitle={The Twelfth International Conference on Learning Representations},
year={2024},
url={https://openreview.net/forum?id=di52zR8xgf}
}

@inproceedings{sd3,
author = {Esser, Patrick and Kulal, Sumith and Blattmann, Andreas and Entezari, Rahim and M\"{u}ller, Jonas and Saini, Harry and Levi, Yam and Lorenz, Dominik and Sauer, Axel and Boesel, Frederic and Podell, Dustin and Dockhorn, Tim and English, Zion and Rombach, Robin},
title = {Scaling rectified flow transformers for high-resolution image synthesis},
year = {2024},
publisher = {JMLR.org},
booktitle = {Proceedings of the 41st International Conference on Machine Learning},
articleno = {503},
numpages = {28},
location = {Vienna, Austria},
series = {ICML'24}
}

@article{storymaker,
  title={Storymaker: Towards holistic consistent characters in text-to-image generation},
  author={Zhou, Zhengguang and Li, Jing and Li, Huaxia and Chen, Nemo and Tang, Xu},
  journal={arXiv preprint arXiv:2409.12576},
  year={2024}
}

@article{storynizor,
  title={Storynizor: Consistent Story Generation via Inter-Frame Synchronized and Shuffled ID Injection},
  author={Ma, Yuhang and Xu, Wenting and Zhao, Chaoyi and Sun, Keqiang and Jin, Qinfeng and Zhao, Zeng and Fan, Changjie and Hu, Zhipeng},
  journal={arXiv preprint arXiv:2409.19624},
  year={2024}
}

@misc{flux,
    author={Black Forest Labs},
    title={FLUX},
    year={2024},
    howpublished={\url{https://github.com/black-forest-labs/flux}},
}

@misc{flux1kontext,
      title={FLUX.1 Kontext: Flow Matching for In-Context Image Generation and Editing in Latent Space}, 
      author={Black Forest Labs and Stephen Batifol and Andreas Blattmann and Frederic Boesel and Saksham Consul and Cyril Diagne and Tim Dockhorn and Jack English and Zion English and Patrick Esser and Sumith Kulal and Kyle Lacey and Yam Levi and Cheng Li and Dominik Lorenz and Jonas Müller and Dustin Podell and Robin Rombach and Harry Saini and Axel Sauer and Luke Smith},
      year={2025},
      eprint={2506.15742},
      archivePrefix={arXiv},
      primaryClass={cs.GR},
      url={https://arxiv.org/abs/2506.15742},
}

@InProceedings{characonsist,
    author    = {Wang, Mengyu and Ding, Henghui and Peng, Jianing and Zhao, Yao and Chen, Yunpeng and Wei, Yunchao},
    title     = {CharaConsist: Fine-Grained Consistent Character Generation},
    booktitle = {Proceedings of the IEEE/CVF International Conference on Computer Vision (ICCV)},
    month     = {October},
    year      = {2025},
    pages     = {16058-16067}
}

@InProceedings{vista,
    author    = {Dong, Sibo and Ismail Shaheen and Maggie Shen and Rupayan Mallick and Sarah Adel Bargal},
    title     = {ViSTA: Visual Storytelling using Multi-modal Adapters for Text-to-Image Diffusion Models},
    booktitle = {Proceedings of the Winter Conference on Applications of Computer Vision (WACV)},
    month     = {March},
    year      = {2026},
    pages     = {}
}

@article{ren2024grounded,
  title={Grounded sam: Assembling open-world models for diverse visual tasks},
  author={Ren, Tianhe and Liu, Shilong and Zeng, Ailing and Lin, Jing and Li, Kunchang and Cao, He and Chen, Jiayu and Huang, Xinyu and Chen, Yukang and Yan, Feng and others},
  journal={arXiv preprint arXiv:2401.14159},
  year={2024}
}

@inproceedings{clip,
    title = "{CLIPS}core: A Reference-free Evaluation Metric for Image Captioning",
    author = "Hessel, Jack  and
      Holtzman, Ari  and
      Forbes, Maxwell  and
      Le Bras, Ronan  and
      Choi, Yejin",
    editor = "Moens, Marie-Francine  and
      Huang, Xuanjing  and
      Specia, Lucia  and
      Yih, Scott Wen-tau",
    booktitle = "Proceedings of the 2021 Conference on Empirical Methods in Natural Language Processing",
    month = nov,
    year = "2021",
    address = "Online and Punta Cana, Dominican Republic",
    publisher = "Association for Computational Linguistics",
    url = "https://aclanthology.org/2021.emnlp-main.595/",
    doi = "10.18653/v1/2021.emnlp-main.595",
    pages = "7514--7528",
}

@article{dino,
title={{DINO}v2: Learning Robust Visual Features without Supervision},
author={Maxime Oquab and Timoth{\'e}e Darcet and Th{\'e}o Moutakanni and Huy V. Vo and Marc Szafraniec and Vasil Khalidov and Pierre Fernandez and Daniel HAZIZA and Francisco Massa and Alaaeldin El-Nouby and Mido Assran and Nicolas Ballas and Wojciech Galuba and Russell Howes and Po-Yao Huang and Shang-Wen Li and Ishan Misra and Michael Rabbat and Vasu Sharma and Gabriel Synnaeve and Hu Xu and Herve Jegou and Julien Mairal and Patrick Labatut and Armand Joulin and Piotr Bojanowski},
journal={Transactions on Machine Learning Research},
issn={2835-8856},
year={2024},
url={https://openreview.net/forum?id=a68SUt6zFt},
note={Featured Certification}
}

@inproceedings{dreamsim,
author = {Fu, Stephanie and Tamir, Netanel Y. and Sundaram, Shobhita and Chai, Lucy and Zhang, Richard and Dekel, Tali and Isola, Phillip},
title = {DreamSim: learning new dimensions of human visual similarity using synthetic data},
year = {2023},
publisher = {Curran Associates Inc.},
address = {Red Hook, NY, USA},
booktitle = {Proceedings of the 37th International Conference on Neural Information Processing Systems},
articleno = {2208},
numpages = {27},
location = {New Orleans, LA, USA},
series = {NIPS '23}
}

@inproceedings{stanza,
 author = {Qi, Peng and Zhang, Yuhao and Zhang, Yuhui and Bolton, Jason and Manning, Christopher D.},
 booktitle = {Proceedings of the 58th Annual Meeting of the Association for Computational Linguistics: System Demonstrations},
 title = {Stanza: A {Python} Natural Language Processing Toolkit for Many Human Languages},
 url = {https://nlp.stanford.edu/pubs/qi2020stanza.pdf},
 year = {2020}
}

@article{ip-adapter,
  title={Ip-adapter: Text compatible image prompt adapter for text-to-image diffusion models},
  author={Ye, Hu and Zhang, Jun and Liu, Sibo and Han, Xiao and Yang, Wei},
  journal={arXiv preprint arXiv:2308.06721},
  year={2023}
}

@ARTICLE{otsu,
  author={Otsu, Nobuyuki},
  journal={IEEE Transactions on Systems, Man, and Cybernetics}, 
  title={A Threshold Selection Method from Gray-Level Histograms}, 
  year={1979},
  volume={9},
  number={1},
  pages={62-66},
  doi={10.1109/TSMC.1979.4310076}}

@inproceedings{
dift,
title={Emergent Correspondence from Image Diffusion},
author={Luming Tang and Menglin Jia and Qianqian Wang and Cheng Perng Phoo and Bharath Hariharan},
booktitle={Thirty-seventh Conference on Neural Information Processing Systems},
year={2023},
url={https://openreview.net/forum?id=ypOiXjdfnU}
}

@inproceedings{
ditft,
title={Unleashing Diffusion Transformers for Visual Correspondence by Modulating Massive Activations},
author={Chaofan Gan and Yuanpeng Tu and Xi Chen and Tieyuan Chen and Yuxi Li and Mehrtash Harandi and Weiyao Lin},
booktitle={The Thirty-ninth Annual Conference on Neural Information Processing Systems},
year={2025},
url={https://openreview.net/forum?id=s3MwCBuqav}
}

\newpage
\appendix

\section{Method}

\subsection{Dynamic Mask Extraction}

\begin{wrapfigure}{r}{0.4\linewidth}
\vspace{-20pt}
\centering
\includegraphics[width=\linewidth]{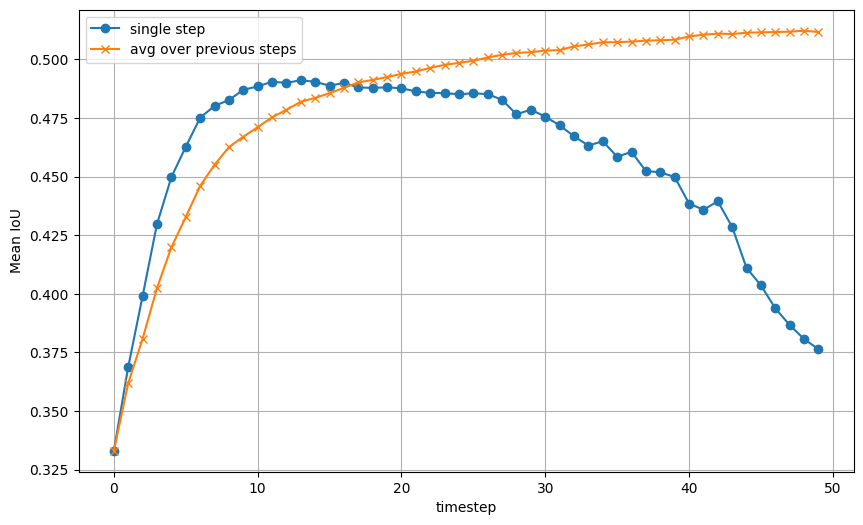}
\caption{Mean IoU between attention-derived masks and Grounded SAM segmentation across diffusion timesteps.}
\label{fig:mask_iou}
\end{wrapfigure}

To better understand mask quality, we analyze the localization accuracy of attention-derived masks across diffusion timesteps.
Specifically, we generate images and obtain ground-truth character masks using Grounded SAM~\cite{ren2024grounded}.
We then compute the Intersection-over-Union (IoU) between the attention-derived masks and the Grounded SAM masks.

As shown in \cref{fig:mask_iou}, during early timesteps the attention map from the current timestep provides better localization accuracy.
In later timesteps, averaging attention maps over previous steps produces more stable and accurate masks.
Based on this observation, we adopt a dynamic mask estimation strategy:
for early timesteps we use the current attention map, while for later timesteps we use the average attention map over previous timesteps.

\subsection{Dataset Construction}
\label{sec:appendix_dataset}
\subsubsection{Data Construction Pipeline}
We employed a ``Dual-Generation'' approach to data synthesis. This pipeline leverages a Large Language Model (LLM) to simultaneously generate a linguistically natural surface realization of a story alongside a strictly parameterized semantic template. This dual approach allows us to maintain natural linguistic flow while affording precise programmatic control over how characters are referenced across the narrative.

\paragraph{Phase 1: Structured Story Generation (JSON).}
We utilized an LLM as a ``Technical Dataset Generator'' to create the core narrative logic. To ensure structural integrity and granular control, the model was heavily prompted to output strict JSON objects containing:

\begin{enumerate}
    \item \textbf{Metadata:} Global story attributes including genre, character count, and thematic keywords.
    \item \textbf{Character Bank:} Detailed attributes for every entity, establishing their category (Human, Animal, Object, Fictional), type (\eg., ``Robot''), proper name, comprehensive pronoun mappings (subject, object, possessive, reflexive), and a detailed visual description.
    \item \textbf{Scene List:} A sequence of 6 scenes per story. Crucially, the action in each scene was generated in two paired formats:
    \begin{itemize}
        \item \texttt{story\_action}: A linguistically sound, coherent natural language description of the scene.
        \item \texttt{semantic\_action}: A programmatic template of the exact same action where all character mentions and pronouns are replaced with explicit structured placeholders (\eg., \texttt{[c1.possessive]}). This also included complex plural mapping identifiers like \texttt{[all.subject]}.
    \end{itemize}
\end{enumerate}

This paired parameterization establishes a perfectly resolved underlying logical state while retaining the contextual grammar of the original text.

\paragraph{Phase 2: Difficulty Stratification (The Rendering Layer).}
Once the semantic templates were validated, we procedurally generated configurations by substituting the placeholders in the \texttt{semantic\_action} based on strict rules. This step introduces the core contribution of our dataset: controlled co-reference and visual consistency difficulty. We mapped the templates into six distinct variations as depicted in \cref{tab:dataset_variants}

\begin{table}[tb]
\footnotesize
    \centering
    \caption{Dataset Variations for Co-reference and Visual Consistency Difficulty.}
    \begin{tabular}{p{4.0cm}|p{8.5cm}}
        \toprule
        \textbf{Configuration Mode} & \textbf{Description} \\
        \midrule
        \textbf{V1: Repeated} & 
        The full visual description replaces the subject in all scenes.\\
        \addlinespace
        \textbf{V2: Single $\rightarrow$ Mix} & 
        Uses the LLM's original \texttt{story\_action}. Description is used once, followed by natural, context-aware pronouns and types.\\
        \addlinespace
        \textbf{V3: Single $\rightarrow$ Type} & 
        Description is given in scene 1. Subsequent mentions strictly use the character's class/type.\\
        \addlinespace
        \textbf{V4: Single $\rightarrow$ Name} & 
        Description is given in scene 1. Subsequent mentions strictly use the character's arbitrary Proper Name.\\
        \addlinespace
        \textbf{V5: No Desc $\rightarrow$ Type} & 
        Visual description is omitted entirely; the subject is exclusively referred to by its generic type across all scenes.\\
        \addlinespace
        \textbf{V6: No Desc $\rightarrow$ Name} & 
        Visual description is omitted entirely; the subject is exclusively referred to by its arbitrary Proper Name.\\
        \bottomrule
    \end{tabular}
    \label{tab:dataset_variants}
\end{table}
The different variations correspond to different difficulty levels and depending on the task, they measure different abilities. For example: V1 can be used to measure a model's ability to handle visual character consistency given sufficient textual cues, while V5 measures the same ability but only given visual cues from previous frame. On the other hand, comparing a model's performance on V1 \vs V2 can give insight on the model's ability to handle textual co-reference resolution.

\paragraph{Statistics, Composition, and Quality Assurance.}
The dataset consists of 200 unique narrative logic flows (1,200 generated variations), split evenly across Single-Character (100) and Dual-Character (100) stories. Characters are distributed equally across four taxonomic categories: Human, Animal, Object, and Fictional.
To handle LLM hallucination, a schema validation step was employed to reject any JSON output that failed to meet the strict structural schema.

\subsubsection{LLM Prompt}
To generate the structured semantic pairs, we utilized the following one-shot system prompt:
\begin{lstlisting}[style=markdown]
You are a technical dataset generator. Your sole purpose is to create story entries for a visual storytelling dataset.
You MUST respond with a single, valid JSON object.
Do NOT provide any conversational text, markdown formatting like ````json`, or any explanation. Your entire response must be the JSON object itself. The reply will be parsed automatically.
Your task is to generate one new story entry. This entry must strictly follow the schema, example, and generation rules provided.

1. JSON Schema
Your output must validate against this schema:
```json
[
  {
    "story_id": 1, // Integer: Unique ID for the story. You will be given this in the task section at the end.
    "metadata": {
      "title": "Story Title", // String: A human-readable title.
      "genre": ["Genre"], // List[String]: \eg., "Sci-Fi", "Fantasy", "Realism".
      "num_scenes": 3, // Integer: Total number of scenes. You will be given this in the task section at the end.
      "num_characters": 1, // Integer: Total number of characters. You will be given this in the task section at the end.
      "keywords": ["key", "words"] // List[String]: Keywords for the story.
    },
    "characters": [
      {
        "char_id": "c1", // String: Unique character ID (c1, c2, etc.).
        "category": "object", // String: One of (human, animal, object, fictional). This is the main character category. You will be given this in the task section at the end.
        "name": "Beeper", // String: Character's proper name (or "" if none).
        "type": "Robot", // String: Character's class/type (\eg., "Girl", "Cat", "Robot", "Fairy").
        "subject": "it", // String: Subject pronoun (he, she, it).
        "object": "it", // String: Object pronoun (him, her, it).
        "possessive_adv": "its", // String: Possessive adjective (his, her, its).
        "possessive": "its", // String: Possessive pronoun (his, hers, its).
        "reflexive": "itself", // String: Reflexive pronoun (himself, herself, itself).
        "description": "small robot with..." // String: visual description, starting with a lowercase letter.
      }
    ],
    "scenes": [
      {
        "scene_id": "s1", // String: Unique scene ID (s1, s2, etc.).
        "background_id": "b1", // String: Unique background ID (b1, b2, etc.). If the background is the same as a previous scene, reuse the same background_id.
        "background": "A sunny park bench...", // String: Full description of the setting used for evaluation. if subsequent scenes happen in the same setting, you can copy the background. Background can stay the same for the entire storyline or change.
        "characters_in_scene": ["c1"], // List[String]: char_ids present in the scene.
        "story_action": "The small robot with... sits on the bench..." // String: story action. Only a single action per scene. More info on this in the generation rules below.
        "semantic_action": "[c1] sits on the bench..." // String: Templated story action. Replace character mentions with character placeholders. More info on this in the generation rules below.
      }
    ]
  }
]
```

2. High-Quality Example
Here is a perfect example of a single story entry.
```json
[
  {
    "story_id": 1,
    "metadata": {
      "title": "The Curious Robot",
      "genre": ["Sci-Fi"],
      "num_scenes": 3,
      "num_characters": 1,
      "keywords": ["robot", "park", "butterfly", "curiosity"]
    },
    "characters": [
      {
        "char_id": "c1",
        "category": "object",
        "name": "Beeper",
        "type": "Robot",
        "subject": "it",
        "object": "it",
        "possessive_adv": "its",
        "possessive": "its",
        "reflexive": "itself",
        "description": "polished small silver robot with a big blue optic sensors"
      }
    ],
    "scenes": [
      {
        "scene_id": "s1",
        "background_id": "b1",
        "background": "A sunny park bench with green grass and distant trees.",
        "characters_in_scene": ["c1"],
        "story_action": "A polished small silver robot with a big blue optic sensors sits on the bench, tilting its head as it watches the sky."
        "semantic_action": "[c1] sits on the bench, tilting [c1.possessive] head as [c1.subject] watches the sky."
      },
      {
        "scene_id": "s2",
        "background_id": "b1",
        "background": "A sunny park bench with green grass and distant trees.",
        "characters_in_scene": ["c1"],
        "story_action": "It stops as a blue butterfly flutters down and lands on the hand of the robot."
        "semantic_action": "[c1] stops as a blue butterfly flutters down and lands on the hand of [c1.type]."
      },
      {
        "scene_id": "s3",
        "background_id": "b2",
        "background": "A wooden house with metal accents with the sun setting, casting a warm glow.",
        "characters_in_scene": ["c1"],
        "story_action": "The robot arrives at its house after a long day, tired and ready to recharge."
        "semantic_action": "[c1] arrives at [c1.possessive] house after a long day, tired and ready to recharge."
      }
    ]
  }
]
```

3. Core Generation Rules
For story_action:
- story_action should be a linguisitically sound and coherent description of the story scene, while semantic_action should be a templated version of the story_action that replaces character mentions with character placeholders.
- The first story_action in the first scene should introduce the character(s) using the character `description` from the `characters` section. Subsequent story_actions should not repeat the character description but can refer to the character using `subject` pronoun or `type` but not `name` from the `characters` section depending on the linuistic contxt and grammer. For example, if the pronoun is ambiguous, use the `type`, if not, use either.
- Other secondary mentions of the character in the same scene should use the appropriate pronouns (`subject`, `object`, `possessive_adv`, `possessive`, `reflexive`) or `type` to avoid ambiguity.
- the `name` attribute in the `characters` section should not be used, it is only for reference.
For semantic_action:
- in semantic_action, main character mentions like the one containing the description in the first scene or the main mentions from the second scene onward should use the `[char_id]` placeholder only even if the mention in story_action is `[char_id.subject]` or`[char_id.type]`. There has to be one instance of `[char_id]` in each semantic_action.
- Other secondary mentions in the same scene should use the appropriate pronoun (`[char_id.subject]`, `[char_id.object]`, `[char_id.possessive_adv]`, `[char_id.possessive]`, `[char_id.reflexive]`) or `[char_id.type]`  placeholders to avoid ambiguity.
- to replace plural references in semantic_action to all characters like "they" or "them", use the placeholder `[all]` with the main mention and with the appropriate pronoun placeholder like `[all.subject]` or `[all.object]` in secondary mentions depending on the linguistic context and grammar.
- In a specific scene, if the characters are not mentioned separately in the story_action and are referred to as a group using a plural pronoun, there should be at least one instance of `[all]` in the semantic_action even if it is the only mention.
- Don't replace references to other non-character entities in the story_action with placeholders in semantic_action, only character mentions should be replaced with placeholders.
Other rules:
- Each story should be a list of scenes that form a naturally flowing story.
- Each scene should contain at most one action per character that drives the story forward.
- Backgrounds can be reused across scenes. If a background is reused, the `background_id` should be the same, and the `background` description should be identical.
- Character Description shouldn't be verbose, 2 clauses maximum. It should start with a lowercase letter.
- The number of characters in the story must match `num_characters` in the metadata.
- The number of scenes must match `num_scenes` in the metadata.

4. TASK
Generate a new, unique story entry as a single JSON object. Constraints for this story that you should follow are:

- story\_id: <story_id>
- num\_characters: <character_count>
- Each scene must include <character_presence> from the `characters` list.
- category: <character_category>
- num\_scenes: 6

Respond ONLY with the single JSON object.
\end{lstlisting}

\subsubsection{Story Example}
\paragraph{Raw JSON.}
The following is an unedited JSON output generated by the LLM, demonstrating the decoupling of \texttt{story\_action} and \texttt{semantic\_action}.
\begin{lstlisting}[
    language=json,
]
{
    "story_id": 1,
    "metadata": {
        "title": "The Archer and the Frost Wolf",
        "genre": [
            "Fantasy"
        ],
        "num_scenes": 6,
        "num_characters": 2,
        "keywords": [
            "archer",
            "wolf",
            "forest",
            "ruins",
            "adventure"
        ]
    },
    "characters": [
        {
            "char_id": "c1",
            "category": "human",
            "name": "Eara",
            "type": "Archer",
            "subject": "she",
            "object": "her",
            "possessive_adv": "her",
            "possessive": "hers",
            "reflexive": "herself",
            "description": "nimble female archer wearing a green tunic and leather bracers"
        },
        {
            "char_id": "c2",
            "category": "animal",
            "name": "Silver",
            "type": "Wolf",
            "subject": "he",
            "object": "him",
            "possessive_adv": "his",
            "possessive": "his",
            "reflexive": "himself",
            "description": "massive grey wolf with a distinctive white patch on his chest"
        }
    ],
    "scenes": [
        {
            "scene_id": "s1",
            "background_id": "b1",
            "background": "A dense forest edge with tall pine trees and a narrow dirt path.",
            "characters_in_scene": [
                "c1",
                "c2"
            ],
            "story_action": "A nimble female archer wearing a green tunic and leather bracers stands at the forest edge while a massive grey wolf with a distinctive white patch on his chest sniffs the ground nearby.",
            "semantic_action": "[c1] stands at the forest edge while [c2] sniffs the ground nearby."
        },
        {
            "scene_id": "s2",
            "background_id": "b1",
            "background": "A dense forest edge with tall pine trees and a narrow dirt path.",
            "characters_in_scene": [
                "c1",
                "c2"
            ],
            "story_action": "She begins to walk down the path, and the wolf follows closely behind her.",
            "semantic_action": "[c1] begins to walk down the path, and [c2] follows closely behind [c1.object]."
        },
        {
            "scene_id": "s3",
            "background_id": "b2",
            "background": "A shallow, rocky river with clear water flowing over smooth stones.",
            "characters_in_scene": [
                "c1",
                "c2"
            ],
            "story_action": "The archer leaps across the wet stones as the wolf splashes through the water to stay by her side.",
            "semantic_action": "[c1] leaps across the wet stones as [c2] splashes through the water to stay by [c1.possessive_adv] side."
        },
        {
            "scene_id": "s4",
            "background_id": "b3",
            "background": "Overgrown stone ruins of an ancient tower surrounded by thick ivy.",
            "characters_in_scene": [
                "c1",
                "c2"
            ],
            "story_action": "They both reach the ruins and scan the area for any signs of movement.",
            "semantic_action": "[all] reach the ruins and scan the area for any signs of movement."
        },
        {
            "scene_id": "s5",
            "background_id": "b3",
            "background": "Overgrown stone ruins of an ancient tower surrounded by thick ivy.",
            "characters_in_scene": [
                "c1",
                "c2"
            ],
            "story_action": "The archer notches an arrow, while the wolf lets out a low growl to protect her.",
            "semantic_action": "[c1] notches an arrow, while [c2] lets out a low growl to protect [c1.object]."
        },
        {
            "scene_id": "s6",
            "background_id": "b4",
            "background": "A high mountain cliff overlooking a valley under a vibrant orange sunset.",
            "characters_in_scene": [
                "c1",
                "c2"
            ],
            "story_action": "Finally, she sits on the cliff's edge with the wolf as they watch the sun disappear.",
            "semantic_action": "Finally, [c1] sits on the cliff's edge with [c2] as [all.subject] watch the sun disappear."
        }
    ]
}
\end{lstlisting}

\paragraph{Rendered Variations.}
Below are the six generated textual variations derived from the \texttt{semantic\_action} templates in the example above. These represent the stratified difficulty tiers used during model evaluation.

\noindent\textbf{V1: Repeated Description} \texttt{(\{`mode': `repeated'\})}
\begin{enumerate}[nosep]
    \item A nimble female archer wearing a green tunic and leather bracers stands at the forest edge while a massive grey wolf with a distinctive white patch on his chest sniffs the ground nearby.
    \item The nimble female archer wearing a green tunic and leather bracers begins to walk down the path, and The massive grey wolf with a distinctive white patch on his chest follows closely behind her.
    \item The nimble female archer wearing a green tunic and leather bracers leaps across the wet stones as The massive grey wolf with a distinctive white patch on his chest splashes through the water to stay by her side.
    \item The nimble female archer wearing a green tunic and leather bracers and The massive grey wolf with a distinctive white patch on his chest reach the ruins and scan the area for any signs of movement.
    \item The nimble female archer wearing a green tunic and leather bracers notches an arrow, while The massive grey wolf with a distinctive white patch on his chest lets out a low growl to protect her.
    \item Finally, The nimble female archer wearing a green tunic and leather bracers sits on the cliff's edge with The massive grey wolf with a distinctive white patch on his chest as they watch the sun disappear.
\end{enumerate}

\noindent\textbf{V2: Single $\rightarrow$ Mix} \texttt{(\{`mode': `single', `fallback': `mix'\})}
\begin{enumerate}[nosep]
    \item A nimble female archer wearing a green tunic and leather bracers stands at the forest edge while a massive grey wolf with a distinctive white patch on his chest sniffs the ground nearby.
    \item She begins to walk down the path, and the wolf follows closely behind her.
    \item The archer leaps across the wet stones as the wolf splashes through the water to stay by her side.
    \item They both reach the ruins and scan the area for any signs of movement.
    \item The archer notches an arrow, while the wolf lets out a low growl to protect her.
    \item Finally, she sits on the cliff's edge with the wolf as they watch the sun disappear.
\end{enumerate}

\noindent\textbf{V3: Single $\rightarrow$ Type} \texttt{(\{`mode': `single', `fallback': `type'\})}
\begin{enumerate}[nosep]
    \item A nimble female archer wearing a green tunic and leather bracers stands at the forest edge while a massive grey wolf with a distinctive white patch on his chest sniffs the ground nearby.
    \item The Archer begins to walk down the path, and The Wolf follows closely behind her.
    \item The Archer leaps across the wet stones as The Wolf splashes through the water to stay by her side.
    \item The Archer and The Wolf reach the ruins and scan the area for any signs of movement.
    \item The Archer notches an arrow, while The Wolf lets out a low growl to protect her.
    \item Finally, The Archer sits on the cliff's edge with The Wolf as they watch the sun disappear.
\end{enumerate}

\noindent\textbf{V4: Single $\rightarrow$ Name} \texttt{(\{`mode': `single', `fallback': `name'\})}
\begin{enumerate}[nosep]
    \item Eara, The nimble female archer wearing a green tunic and leather bracers, stands at the forest edge while Silver, The massive grey wolf with a distinctive white patch on his chest, sniffs the ground nearby.
    \item Eara begins to walk down the path, and Silver follows closely behind her.
    \item Eara leaps across the wet stones as Silver splashes through the water to stay by her side.
    \item Eara and Silver reach the ruins and scan the area for any signs of movement.
    \item Eara notches an arrow, while Silver lets out a low growl to protect her.
    \item Finally, Eara sits on the cliff's edge with Silver as they watch the sun disappear.
\end{enumerate}

\noindent\textbf{V5: No Description $\rightarrow$ Type} \texttt{(\{`mode': `no\_desc', `fallback': `type'\})}
\begin{enumerate}[nosep]
    \item The Archer stands at the forest edge while The Wolf sniffs the ground nearby.
    \item The Archer begins to walk down the path, and The Wolf follows closely behind her.
    \item The Archer leaps across the wet stones as The Wolf splashes through the water to stay by her side.
    \item The Archer and The Wolf reach the ruins and scan the area for any signs of movement.
    \item The Archer notches an arrow, while The Wolf lets out a low growl to protect her.
    \item Finally, The Archer sits on the cliff's edge with The Wolf as they watch the sun disappear.
\end{enumerate}

\noindent\textbf{V6: No Description $\rightarrow$ Name} \texttt{(\{`mode': `no\_desc', `fallback': `name'\})}
\begin{enumerate}[nosep]
    \item Eara stands at the forest edge while Silver sniffs the ground nearby.
    \item Eara begins to walk down the path, and Silver follows closely behind her.
    \item Eara leaps across the wet stones as Silver splashes through the water to stay by her side.
    \item Eara and Silver reach the ruins and scan the area for any signs of movement.
    \item Eara notches an arrow, while Silver lets out a low growl to protect her.
    \item Finally, Eara sits on the cliff's edge with Silver as they watch the sun disappear.
\end{enumerate}

\section{Implementation Details and Additional Results}

\begin{wrapfigure}{r}{0.4\linewidth}
\vspace{-20pt}
\centering
\includegraphics[width=\linewidth]{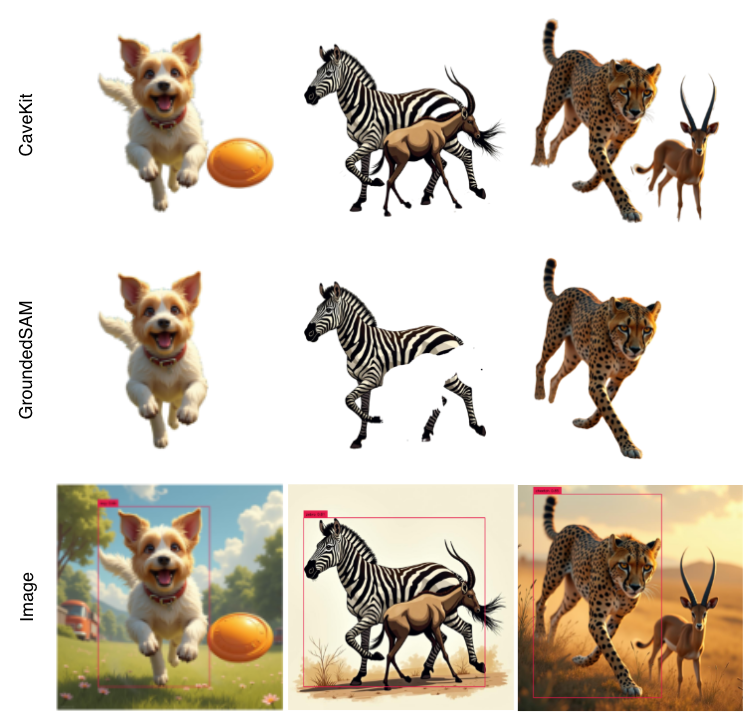}
\caption{Comparison of background removal using CarveKit and Grounded-SAM. 
The examples illustrate three typical failure modes of CarveKit. }
\label{fig:cavekit}
\end{wrapfigure}

\subsection{Background Removal for Evaluation}

Previous works commonly adopt CarveKit for background removal during evaluation. 
However, we find that it is not sufficiently robust for our setting, particularly in images containing multiple characters or object interactions. 
Therefore, we instead employ Grounded-SAM to obtain more reliable foreground segmentation.
\Cref{fig:cavekit} compares the outputs of CarveKit and Grounded-SAM on several representative examples. 
We observe three typical failure modes of CarveKit:
(1) Incorrect foreground detection. CarveKit may include irrelevant objects in the foreground mask (\eg., the flying disc with the dog), leading to inaccurate segmentation.
(2) Failure under occlusion. When objects overlap, CarveKit often retains the occluding object instead of isolating the target instance.
(3) No multi-instance separation. CarveKit cannot separate multiple characters and tends to merge them into a single foreground region.
In contrast, Grounded-SAM produces more accurate instance-level masks, allowing us to reliably isolate the target characters. 
This leads to cleaner foreground extraction and more stable evaluation results.

\subsection{Implementation Details }
We use the FLUX.1-dev as base model and build our method within the FLUX single-stream diffusion transformer blocks. We do not perform any finetuning or parameter modification. All images are generated at a resolution of $1024 \times 1024$ using $50$ diffusion timesteps with classifier-free guidance scale set to $3.5$, and inference is performed in bfloat16 precision. Character mask extraction, correspondence matching, key-value injection, and query blending are applied only within a timestep window $[T_s, T_e]$, where we set $T_s = \text{1}$ and $T_e = \text{30}$. For dynamic mask estimation, the timestep threshold is set to $T_{mask} = \text{15}$. For the dynamic correspondence matching, the threshold is set to $T_{corr} = \text{8}$. The query blending coefficient is fixed to $\lambda = \text{0.2}$ in all experiments. For entity grounding, we use the pretrained English coreference resolution model provided in Stanza~\cite{stanza} to automatically recover the character description token indices from the character-defining prompt. All experiments are conducted on a single NVIDIA L40S GPU.

\subsection{Ablation Study}

\begin{table}[tb]
\caption{
Ablation study on the ConsiStory+ dataset.
Removing query blending degrades character consistency, while further removing dynamic mask extraction and correspondence matching leads to a larger performance drop.
These results demonstrate that both components contribute to consistent character generation.
}\label{tab:ablation}
    \centering
    \begin{tabular}{l|r|r|r|r}
    \toprule
    \textbf{Method} & \multicolumn{1}{l|}{\textbf{CLIP-T} $\uparrow$} & \multicolumn{1}{l|}{\textbf{CLIP-I} $\uparrow$} & \multicolumn{1}{l|}{\textbf{DINO} $\uparrow$} & \multicolumn{1}{l}{\textbf{DreamSim} $\downarrow$} \\
    \midrule
Ours & 0.8369 & \textbf{0.8979} & \textbf{0.7118} & \textbf{0.2234} \\
w/o Query Blending & \textbf{0.8384} & 0.8843 & 0.6717 & 0.2569 \\
w/o Dynamic Mask & \textbf{0.8384} & 0.8833 & 0.6611 & 0.2624 \\
    \bottomrule
    \end{tabular}
\end{table}
\begin{figure}[tb]
    \centering
    \includegraphics[width=\linewidth]{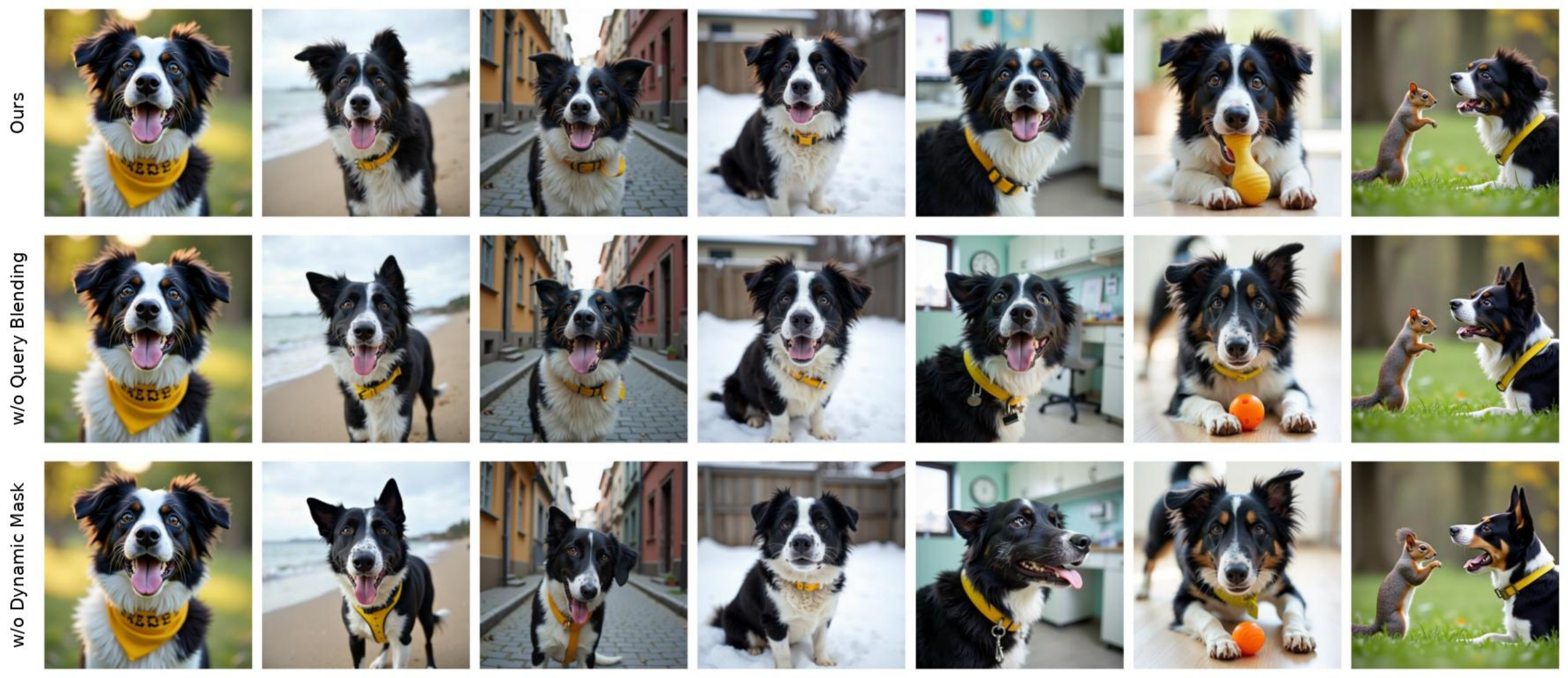}
\caption{
Qualitative ablation results.
Removing query blending reduces character consistency, while removing dynamic mask extraction further leads to identity drift.
}    \label{fig:ablation_vis}
\end{figure}

We conduct an ablation study to analyze the contribution of the proposed components.
\Cref{tab:ablation} reports results starting from the full model and progressively removing modules.
Removing query blending leads to a noticeable drop in character consistency across CLIP-I, DINO, and DreamSim, while having minimal impact on text-image alignment (CLIP-T). 
This indicates that blending query features helps guide attention toward consistent character regions.
When we further remove dynamic mask extraction and correspondence matching, the performance decreases further. This result highlights the importance of accurate character localization and cross-frame correspondence for reliable feature reuse.
Overall, both dynamic mask estimation and query blending play complementary roles in maintaining character consistency across frames.

We further show a qualitative comparison in \cref{fig:ablation_vis}.
Without query blending, the generated character becomes less consistent across frames.
When dynamic mask extraction is also removed, the model often attends to incorrect regions, leading to noticeable identity drift and inconsistent appearance.
These observations align with the quantitative results in \cref{tab:ablation}.

\subsection{Additional Results on Structured Storytelling Benchmark}
\begin{figure}
    \centering
    \includegraphics[width=\linewidth]{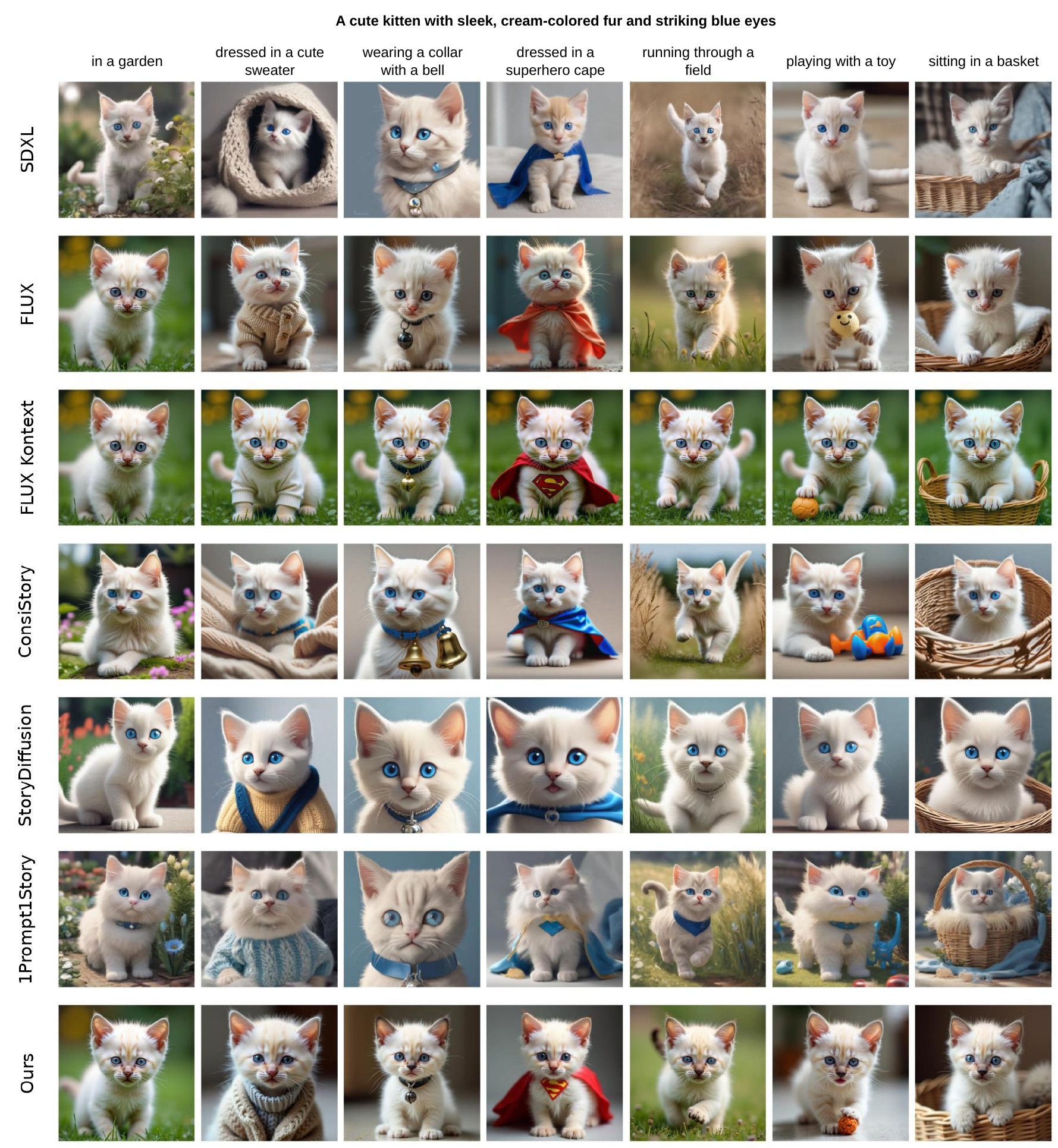}
    \caption{Qualitative comparison results on ConsiStory+ dataset. }
    \label{fig:13_animals_kitten}
\end{figure}

\begin{figure}
    \centering
    \includegraphics[width=0.8\linewidth]{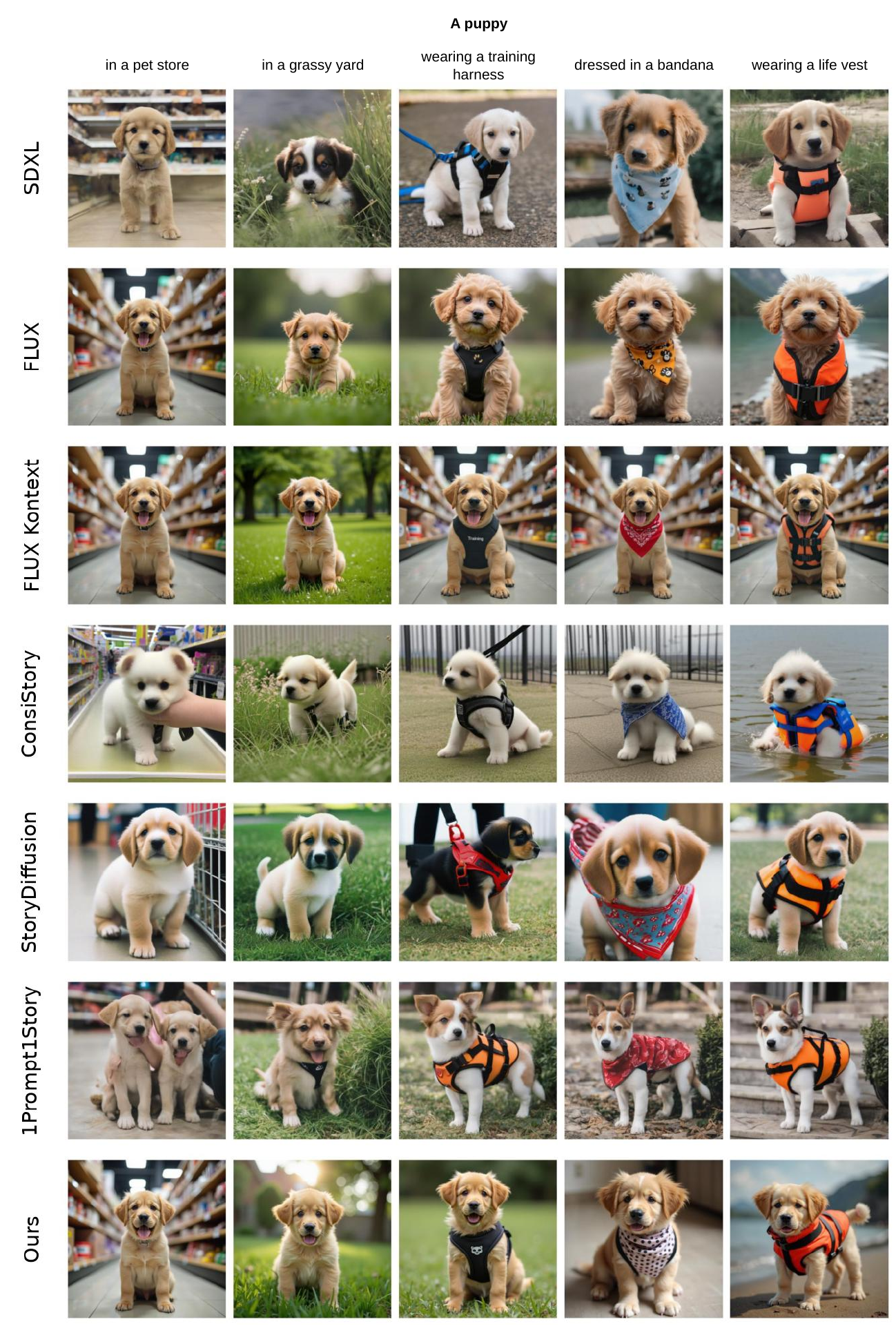}
    \caption{Qualitative comparison results on ConsiStory+ dataset. }
    \label{fig:14_animals_puppy}
\end{figure}

\begin{figure}
    \centering
    \includegraphics[width=0.8\linewidth]{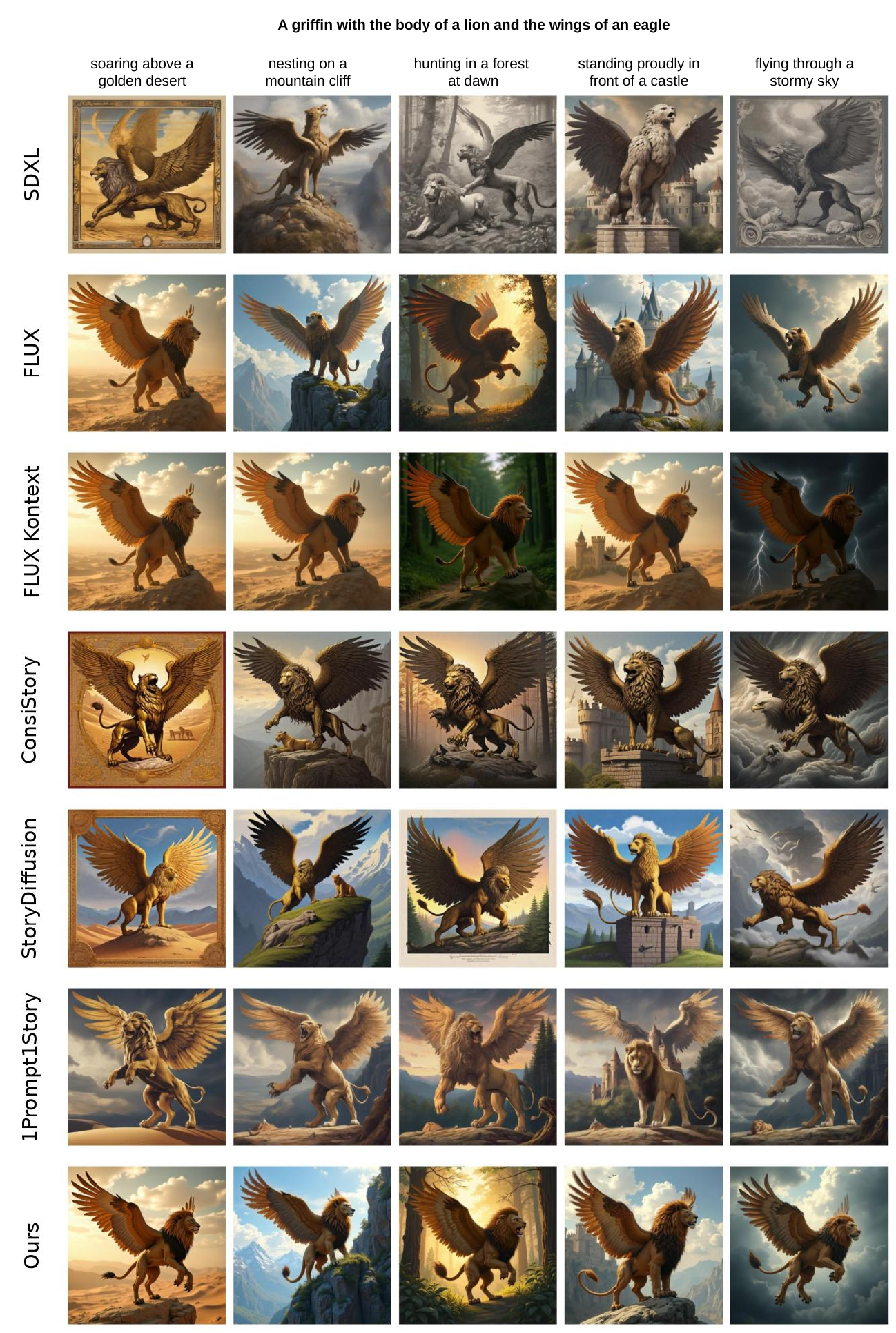}
    \caption{Qualitative comparison results on ConsiStory+ dataset. }
    \label{fig:61_fantasy_griffin}
\end{figure}

\begin{figure}
    \centering
    \includegraphics[width=\linewidth]{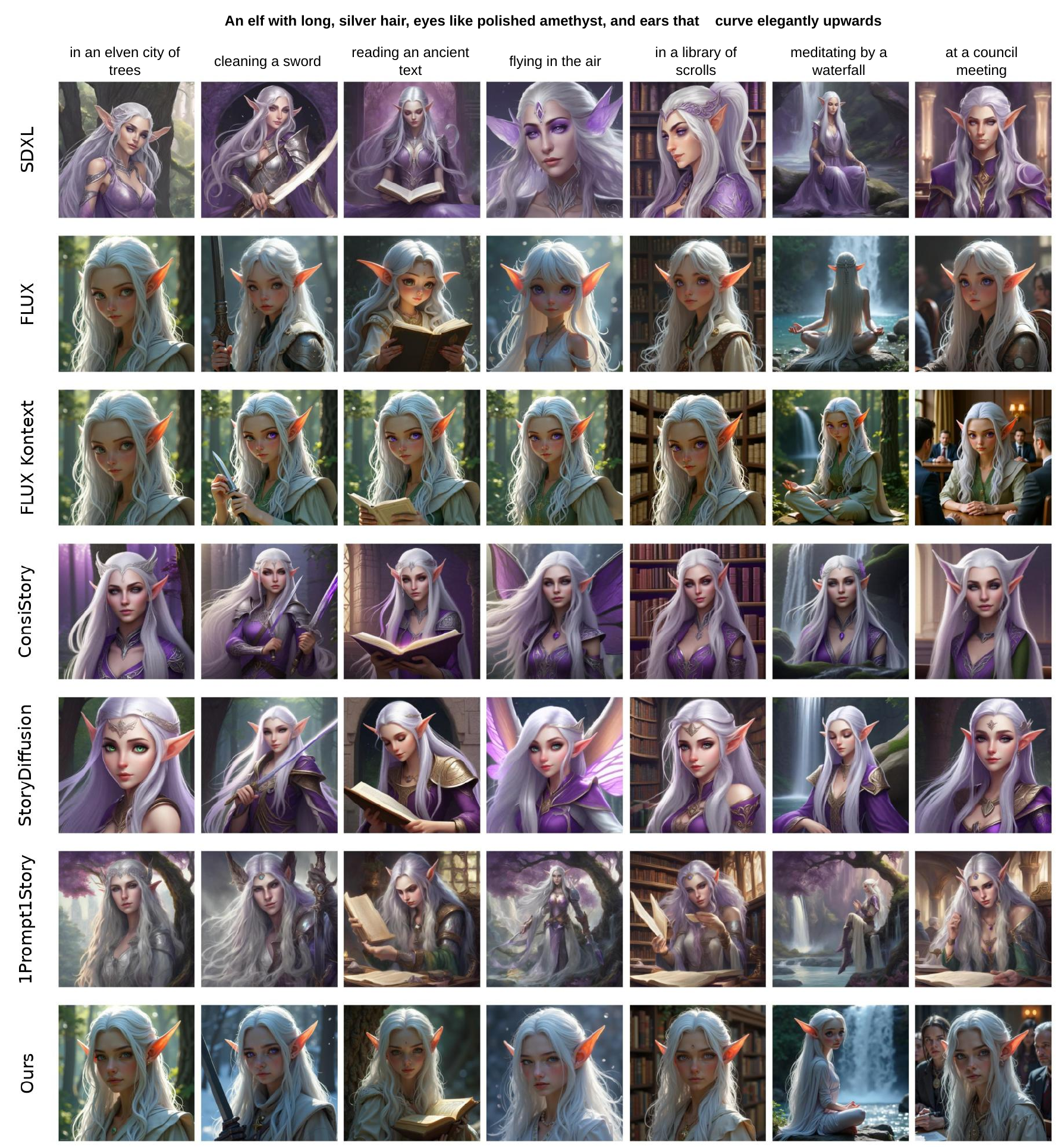}
    \caption{Qualitative comparison results on ConsiStory+ dataset. }
    \label{fig:76_fantasy_elf}
\end{figure}

\cref{fig:13_animals_kitten}, \cref{fig:14_animals_puppy}, \cref{fig:61_fantasy_griffin} and \cref{fig:76_fantasy_elf} show additional qualitative comparison on the ConsiStory+ dataset. 

\begin{figure}
    \centering
    \includegraphics[width=\linewidth]{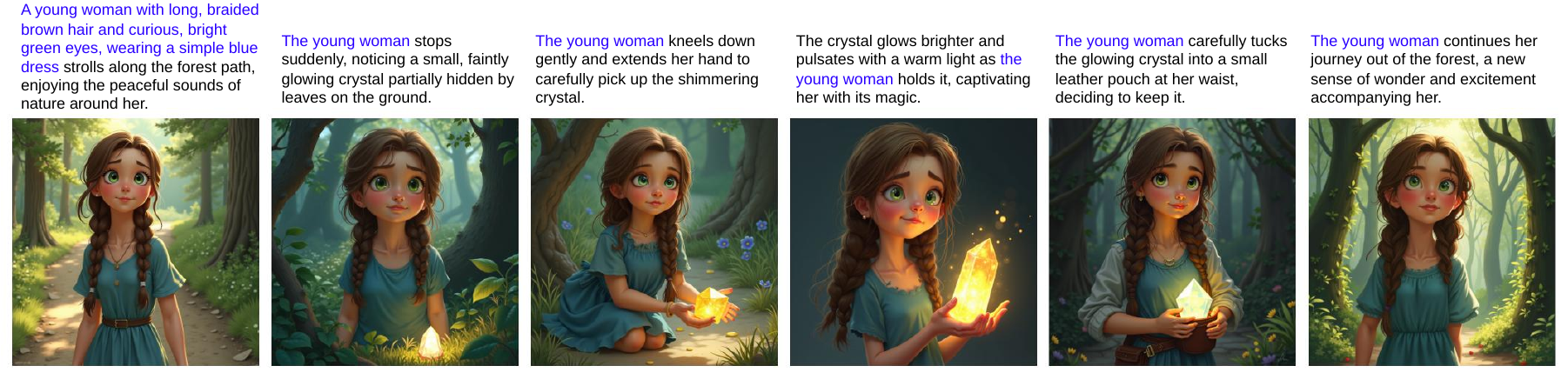}
    \caption{Qualitative results on the FreeStoryBench dataset under the Type setting. Blue text indicates the entity grounding results.}
    \label{fig:12_c1_human_Young}
\end{figure}

\subsection{Additional Results and Discussion on FreeStoryBench}

\begin{figure}
    \centering
    \includegraphics[width=0.95\linewidth]{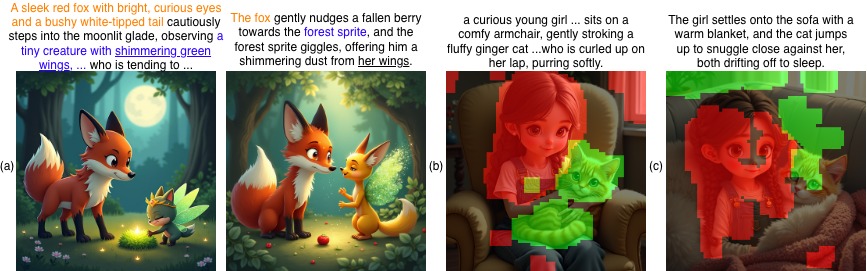}
    \caption{\textbf{Grounding and complex-interaction examples.}
(a) Entity-grounding failure: Stanza incorrectly identifies \underline{her wings} as the second character rather than the \textcolor{blue}{forest sprite}; consequently, only the \textcolor{orange}{fox} remains consistent.
(b) A successful case involving overlapping or occluded character interactions.
(c) A failure case in which inaccurate attention masks lead to localization or consistency errors.}
    \label{fig:fail}
\end{figure}

\paragraph{Qualitative Results.}
\Cref{fig:12_c1_human_Young} presents qualitative results on our proposed FreeStoryBench dataset under the \textbf{Single $\rightarrow$ Type} configuration, where the character-defining prompt contains the full description and subsequent prompts refer to the character using type-based mentions.
The examples demonstrate that our method maintains character consistency even when referring prompts contain only implicit references rather than repeated descriptions.
These results illustrate the effectiveness of our entity grounding and attention-based feature reuse mechanisms under free-form storytelling conditions.

\paragraph{Entity Grounding Analysis.}
We evaluate the accuracy of our Stanza-based entity grounding module across different FreeStoryBench configurations.
For single-character stories, it achieves grounding accuracies of 94.4\% and 97.9\% under the type-based and mixed-reference settings, respectively.
For multi-character stories, the corresponding accuracies are 90.2\% and 93.7\%.
The lower accuracy in the multi-character setting is primarily caused by the increased ambiguity introduced by multiple entities and reference expressions.

In most failure cases, the grounding module fails to associate a referring mention with its corresponding character-description span.
Consequently, the relevant keys and values cannot be reused, and generation for that character effectively falls back to the behavior of the underlying diffusion model.
As shown in \cref{fig:fail}(a), Stanza incorrectly identifies the phrase \emph{her wings} as the second character instead of the \emph{forest sprite}; as a result, only the fox remains visually consistent.

\paragraph{Complex Character Interactions.}
Our method mitigates cross-character interference through entity-aware masks, which constrain feature reuse to the spatial region associated with each character.
As illustrated in \cref{fig:fail}(b), this design can preserve reasonable character consistency even when characters overlap, interact closely, or partially occlude one another.
Nevertheless, such scenarios remain challenging because overlapping or diffuse attention distributions can produce inaccurate or entangled character masks.
This may lead to incorrect localization, feature leakage between characters, or degraded identity consistency, as shown in \cref{fig:fail}(c).

\paragraph{Generalization Beyond Characters.}
Although our entity grounding module is primarily designed to resolve character references, the same mechanism can naturally extend to other entities that should remain visually consistent throughout a story.
For example, objects such as a \emph{basket} may recur across prompts even when the primary characters are two people.
By grounding these object mentions to their corresponding description spans, the model can reuse attention features to maintain a consistent appearance across frames.
This suggests that entity grounding provides a more general mechanism for preserving visual consistency beyond characters, enabling object-level consistency in more complex narratives.

\paragraph{FreeStoryBench Configurations.}
FreeStoryBench supports multiple prompt configurations with different levels of coreference difficulty, as summarized in \cref{tab:dataset_variants}.
In the main paper, we report quantitative results under the \textbf{Single $\rightarrow$ Type} configuration.
This setting represents a canonical free-form storytelling scenario in which the character-defining prompt contains the full description and subsequent prompts refer to the character using type-based mentions.
It serves as a controlled first step toward free-form storytelling by relaxing the structured prompt format used in prior work while retaining a clear reference mechanism.

For evaluation, we assume that all characters appear in the character-defining frame.
This protocol enables a controlled and meaningful evaluation of long-range character consistency, since each character can be tracked across the full story.
Characters introduced only in later frames would provide fewer observations and make consistency scores less directly comparable across stories.
We therefore adopt this setting as an evaluation design choice while leaving more dynamic storytelling scenarios with an evolving cast for future study.

We exclude the \textbf{Repeated} configuration from the main evaluation because it corresponds to the structured prompt format used in prior storytelling benchmarks and has already been extensively studied.
The remaining configurations introduce more varied reference expressions, including mixed reference types, and therefore increase the linguistic complexity of the task.
These settings can be viewed as extensions of the same free-form formulation.
We therefore adopt the \textbf{Single $\rightarrow$ Type} configuration as the primary benchmark for evaluating free-form visual storytelling.

\section{Limitations}

Despite the promising results, our method has several limitations.
First, our approach relies on attention-derived masks to localize characters and constrain feature reuse.
When prompts involve multiple characters, close interactions, or partial occlusions, their attention distributions may become diffuse or overlapping.
This can produce inaccurate or entangled character masks, leading to imperfect localization, cross-character feature leakage, and reduced identity consistency.
Although entity-aware masking mitigates such interference in many cases, complex spatial interactions remain challenging.

Second, the current pipeline is evaluated under a setting in which all characters are introduced in the character-defining frame, and therefore does not cover stories with an evolving cast.
Since the generation module stores and reuses features independently for different entities, the framework can naturally be extended to support characters introduced in later prompts.
In future work, we plan to incorporate dynamic character detection and registration, allowing newly introduced characters to be added to the reusable character memory and consistently maintained in subsequent frames.

Third, the effectiveness of feature reuse depends on the accuracy of the entity grounding module.
Grounding becomes more difficult in multi-character narratives containing ambiguous pronouns, possessive expressions, or multiple semantically similar entities.
When a referring mention is not correctly matched to its character-description span, the corresponding keys and values cannot be reused, and generation for that character effectively falls back to the underlying diffusion model.
More robust grounding models, potentially combining linguistic coreference resolution with visual context accumulated across frames, could reduce these failures.

Finally, our evaluation pipeline relies on automatic segmentation tools to detect and crop characters for consistency measurement.
Although we employ a relatively robust segmentation method, errors may still occur for highly stylized, partially occluded, or fantasy characters.
Such characters often exhibit unusual shapes, exaggerated proportions, or ambiguous semantic categories that are difficult for standard segmentation models to localize.
These errors may introduce noise into the reported evaluation metrics, particularly in complex multi-character scenes.


\end{document}